\documentclass{article}
\usepackage[most]{tcolorbox}
\tcbuselibrary{listings}
\PassOptionsToPackage{numbers, compress}{natbib}
\usepackage[preprint]{neurips_2026}
\usepackage{tabularx}   % for tabularx and X column
\usepackage{graphicx}   % for \resizebox
\usepackage{multirow}   % for \multirow
\usepackage{microtype}
\usepackage{hyperref}
\usepackage{url}
\usepackage{booktabs}
\usepackage{amsmath}
\usepackage{lineno}
\usepackage{comment}
\usepackage{subcaption}
\usepackage{amssymb}% 
\usepackage{pifont}
\newcommand{\cmark}{\ding{51}}%
\newcommand{\xmark}{\ding{55}}%
\newcommand{\gain}[2]{#1$_{\textcolor{mygreen}{\scalebox{1.0}{\tiny +#2}}}$}
\newcommand{\improve}[2]{#1$_{\textcolor{mygreen}{\scalebox{1.0}{\tiny -#2}}}$}
\newcommand{\drop}[2]{#1$_{\textcolor{red}{\scalebox{1.0}{\tiny -#2}}}$}

\usepackage{enumitem}       % For better list control
\usepackage{cleveref}       % For smart referencing (Box 1, etc.)
\newtcolorbox[auto counter, number within=section]{planbox}[2][]{%
    colback=gray!5!white,    % Background color
    colframe=gray!75!black,  % Border color
    fonttitle=\bfseries,     % Bold title
    title={\thetcbcounter: #2}, % Displays "Box 1: Title"
    label={#1},              % This allows the \label to work
    enhanced,                % Allows advanced styling
    breakable                % Allows box to split across pages
}

\newtcolorbox{casebox}[1]{%
    colback=white,
    colframe=gray!20!black,
    fonttitle=\bfseries,
    title={#1},
    enhanced,
    breakable,
    before skip=10pt,
    after skip=10pt
}

\newtcolorbox[use counter from=planbox]{failurebox}[2][]{%
    colback=red!5!white,
    colframe=red!75!black,
    fonttitle=\bfseries,
    title={Box \thetcbcounter: #2},            % Second argument is the title
    label={#1},            % First argument is the label
    enhanced,
    breakable
}

\usepackage{array}      % for \newcolumntype and >{\...}

\usepackage[utf8]{inputenc} % allow utf-8 input
\usepackage[T1]{fontenc}    % use 8-bit T1 fonts
\usepackage{hyperref}       % hyperlinks
\usepackage{url}            % simple URL typesetting
\usepackage{booktabs}       % professional-quality tables
\usepackage{amsfonts}       % blackboard math symbols
\usepackage{nicefrac}       % compact symbols for 1/2, etc.
\usepackage{microtype}      % microtypography
\usepackage[table]{xcolor}         % colors
\usepackage[normalem]{ulem}

\newcommand{\haolun}[1]{{\color{red}{[HaoLun: #1]}}}

\title{Organize then Retrieve: Hierarchical Memory Navigation for Efficient Agents}
\author{
Hao-Lun Hsu$^{1}$, Nikki Lijing Kuang$^{2}$, Boyi Liu$^{2}$, Zhewei Yao$^{2}$, Yuxiong He$^{2}$ \\
$^{1}$Duke University 
$^{2}$Snowflake AI Research
}

\begin{document}
\definecolor{mygreen}{RGB}{0,128,0}
\definecolor{lightblue}{RGB}{220,225,255}
\maketitle

\begin{abstract}
Large language model (LLM) agents struggle with long-horizon tasks due to their inherent statelessness, requiring all task-relevant information to be encoded in growing input contexts. The resulting degraded reasoning quality, increased inference cost, and higher latency necessitate efficient working memory mechanisms. However, existing approaches either rely on lossy compression or similarity-based retrieval, which often fail to capture temporal structure and causal dependencies required for multi-step agentic tasks. In this work, we present HORMA, a \textbf{H}ierarchical \textbf{O}rganize-and-\textbf{R}etrieve \textbf{M}emory \textbf{A}gent that organizes experience into a file-system-like hierarchical structure, where summarized entities are linked to the corresponding raw trajectories, enabling efficient access without losing detailed information. HORMA decomposes working memory into two stages: structured memory construction and navigation-based retrieval. The construction module iteratively refines how experiences are structured by distinguishing between failures caused by missing information and those caused by misleading or overloaded context. The navigation module retrieves task-relevant context by traversing the hierarchy using a lightweight agent  trained with reinforcement learning to select minimal yet sufficient context, thereby reducing latency along the critical execution path. Across ALFWorld, LoCoMo, and LongMemEval, HORMA improves task performance under constrained context budgets while requiring at most 22.17\% of the baseline token usage in long conversation tasks. Compared to existing methods, it consistently achieves better efficiency-performance trade-offs and generalizes effectively to unseen tasks.

\end{abstract}

\section{Introduction}

\begin{figure*}
    \centering 
    \includegraphics[height=0.31\textwidth]{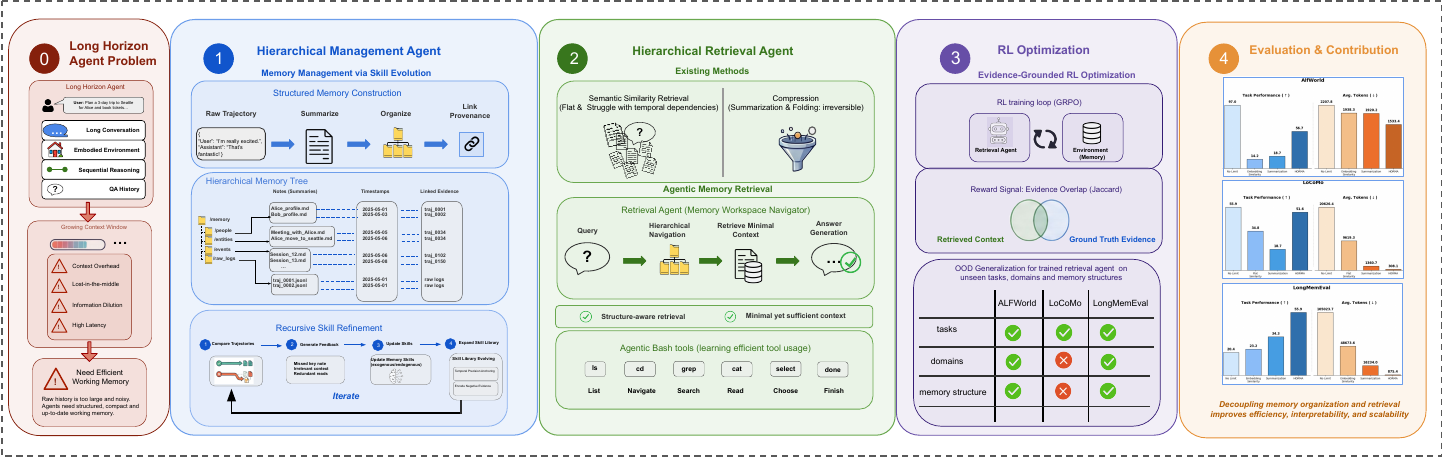}\caption{Overview of HORMA Framework. The system aims to solve the long-horizon problem (0) and explicitly decouples working memory into two specialized modules (1) \& (2), accompanied by its dedicated retrieval training and verification benchmarks (3 \& 4): (1) Hierarchical Management Agent, which organizes raw trajectories into structured, linked notes within a file-system workspace using recursive skill refinement; (2) Hierarchical Retrieval Agent, which navigates this hierarchy using Bash tools and terminal actions to select task-relevant context.
    }\label{fig:main} 
\end{figure*}

 In agentic systems, working memory functions as a short-term workspace that allows the agent to maintain task-relevant information in complex long-horizon tasks. Existing approaches suffer from two key limitations: agents either act as history hoarders (see Figure~\ref{fig:main} (0)), retaining large amounts of history~\citep{wei2022chain, yao2023react}, leading to context overload~\citep{an2025effective}, information dilution~\citep{liu2024lostmiddle}, prohibitive latency and high inference cost~\citep{kang2025acon}, or rely on lossy compression mechanisms~\citep{jiang-etal-2023-llmlingua, jiang-etal-2024-longllmlingua, li2023compressing, yoon2024compact}, including summarization~\citep{lu2025summarizationrl, wang2025recursively, wu2025resum} and context folding~\citep{sun2025contextfolding, ye2025agentfold}, which irreversibly discard fine-grained information necessary for downstream reasoning~\citep{laban2026llms, lampinen2025latent, pan2025secom, ravaut-etal-2024-context, wu2026memory}.
 
 To address these limitations, recent work has delegated working memory to explicit external storage systems~\citep{kang2025memoryos, packer2023memgpt, chhikara2025mem0, xu2025amem, memoryr1,  zhong2024memorybank}. Despite improving storage scalability, existing external memory architectures typically organize experience as flat collections of independent entries retrieved through semantic similarity~\citep{chhikara2025mem0, karpukhin-etal-2020-dense, salama-etal-2025-meminsight, xu2025amem}. Such designs fail to capture temporal hierarchies and causal dependencies accumulated over long interaction horizons. As a result, retrieval often degenerates into shallow semantic matching that surfaces temporally inconsistent or contextually irrelevant information~\citep{zhuang2026linearrag, memoryr1, zhang2026memrl, memoryt1}. Effective long-horizon memory therefore requires not only selective retention, but also hierarchical organization of accumulated experience into reusable and semantically coherent structures~\citep{kang2025memoryos, xu2026structmem, zeng2024structural}, thereby improving downstream task performance.

To support such structured long-horizon memory, most existing memory systems treat memory construction and retrieval as a monolithic system that is jointly optimized within a unified framework~\citep{memoryt1, zhang2026memrl, mem1, yu2026memagent, sun2025contextfolding}. However, memory construction and retrieval serve fundamentally different functional roles and admit distinct optimization strategies. Memory construction determines how experiences are abstracted and structurally organized over time. Its impact often manifests only after extended interactions, making its quality difficult to assess through immediate task outcomes. Furthermore, modern proprietary LLMs already demonstrate strong capabilities for semantic abstraction and hierarchical structuring~\citep{jiang2025hibench, son2026content, jin-etal-2025-disentangling-memory}, suggesting that effective memory structures can often be induced directly from their existing capabilities. %This suggests that effective memory organization can be largely achieved by leveraging the model's existing abstraction abilities, rather than one that must be learned through task-level adaptation. In contrast, memory retrieval is a repeated decision-making process executed at inference time. Retrieval mistakes have immediate and localized effects on downstream reasoning, creating clear optimization signals. Consequently, retrieval is naturally better suited for direct optimization. 
In contrast, memory retrieval determines which information is exposed to the agent at inference time and therefore directly influences downstream decisions. Consequently, retrieval is naturally more amenable to explicit optimization.

This distinction becomes particularly problematic in reinforcement learning (RL)-based memory systems. Jointly optimizing memory construction and retrieval through sparse task-level rewards introduces a severe \textit{credit assignment gap}~\citep{sun2025beyond}: when an agent fails a long-horizon task, it becomes unclear whether the failure originates from poor memory organization, inaccurate retrieval, or downstream reasoning~\citep{yuan2025memsearcher, MemSkill, tan2026hindsight}. As a result, sparse outcome rewards provide weak and entangled supervision signals for both components. Existing attempts to mitigate this issue through intermediate or multi-level rewards partially alleviate the optimization difficulty, but they often require carefully engineered reward designs~\citep{memoryt1, wang2025memalpha} and generalize poorly beyond conversational settings~\citep{memoryt1}.

Motivated by these observations, we propose HORMA, a \textbf{H}ierarchical \textbf{O}rganize-and-\textbf{R}etrieve \textbf{M}emory \textbf{A}gent that explicitly decouples memory construction from retrieval within a shared hierarchical file-system workspace (Figure~\ref{fig:main}). Both modules are implemented as tool-using agents that interact with the workspace through executable file-system operations and Bash tools, while serving distinct functional roles. The memory construction module is responsible for maintaining semantically organized memory structures that provide stable abstractions for long-horizon reasoning. Rather than optimizing memory construction directly through unstable long-horizon RL, HORMA treats memory construction as a continual management skill acquisition process. We initialize a domain-agnostic construction policy using proprietary LLMs with strong hierarchical reasoning capabilities~\citep{jiang2025hibench, son2026content}, and iteratively refine this policy through contrastive analysis between successful and failed trajectories. Over time, the construction module accumulates reusable memory management skills~\citep{anthropic2024claude3} that transfer across tasks without relearning memory construction from scratch.

In contrast, the retrieval module operates directly on the inference path and is responsible for efficiently extracting task-relevant context from the hierarchical workspace. Instead of relying on flat semantic retrieval, the retrieval agent actively navigates the organized memory structure through dedicated Bash tools, enabling more temporally consistent and causally grounded access to historical information~\citep{li2026beyond, xu2026structmem}. We further introduce two executable actions, {\texttt{select}, \texttt{done}}, that allow the agent to iteratively verify retrieved memory and uncover missing contextual details~\citep{yang2026beyond}. To enable retrieval-specific optimization beyond sparse task-level supervision, we introduce an auxiliary learning signal (i.e., evidence-grounded retrieval reward) based on overlap between retrieved context and task-relevant ground-truth evidence. This provides direct, fine-grained feedback on retrieval quality that is decoupled from downstream reasoning performance. Leveraging this signal, we optimize the retrieval policy using RL on a lightweight backbone, enabling efficient context extraction under constrained context budgets while reducing computational overhead.

We evaluate HORMA on three challenging long-horizon benchmarks. On ALFWorld~\citep{alfworld}, HORMA achieves higher success rates under both small and large context limits while improving Pareto efficiency between interaction steps and token usage. On long-conversation benchmarks, HORMA significantly reduces context consumption, using only 3.07\%--22.17\% of the tokens required by different baselines on LoCoMo~\citep{locomo} and 1.24\%--16.19\% on LongMemEval~\citep{wu2025longmemeval}. Notably, the learned lightweight retrieval agent exhibits strong out-of-distribution generalization on LongMemEval, outperforming all baselines, including those without context constraints. Overall, these results demonstrate that explicitly decoupling memory management and retrieval yields a more efficient, interpretable, and scalable mechanism for working memory under strict context limits.

\section{Related Work}
\paragraph{Working Memory in LLM-Based Agents.}
Working memory approaches differ in whether they emphasize compression and structuring prior to context entry or dynamic, policy-driven maintenance during execution, but both aim to mitigate context saturation while preserving task-relevant information for reasoning~\citep{huang2026rethinking}.
One line of work focuses on pre- or in-context state formation, compressing or restructuring interaction history before or as it enters the active context. Methods such as ReSum~\citep{wu2025resum} and ACON~\citep{kang2025acon} perform learned compression of trajectories into compact reasoning states, while hierarchical folding~\citep{sun2025contextfolding, ye2025agentfold} and subgoal-based methods~\citep{hu2025hiagent, wang2026subgoal} introduce restructuring to organize long-horizon interactions into manageable abstractions.
A second line of work addresses online maintenance of working memory during execution, directly operating on the evolving context under fixed budgets. Approaches~\citep{yu2026memagent, yuan2025memsearcher, zhang2025memoryaction} use recurrent updates to maintain compact states, while policy-based methods~\citep{chhikara2025mem0, memoryr1} treat memory operations as actions that decide what to store, update, or discard during interaction.

\paragraph{RL for LLMs.}
Reinforcement learning (RL) has become a core technique for improving performance in LLMs~\citep{ppo, grpo, dpo}. RL enables the emergence of reasoning-centric models such as DeepSeek-R1~\citep{guo2025deepseek} and Search-R1~\citep{jin2025searchr}. However, these approaches typically rely on retaining the entire interaction trajectory, leading to scalability and efficiency limitations in long-horizon settings. Recent work has begun exploring memory construction and management through RL. Early approaches~\citep{mem1, yu2026memagent} train models to maintain lightweight text-based memories. Subsequent methods introduce richer memory representations together with simplified memory tool interfaces~\citep{memoryr1, zhang2025learn, zhang2023large}. In contrast to prior end-to-end memory-augmented RL approaches, we formulate memory retrieval as a navigation problem and purely train a dedicated retrieval agent with RL, mitigating the credit assignment challenges.

\paragraph{Memory and Skill Evolution.}
The ability to abstract complex experiences into reusable skills is fundamental to self-improving agents~\citep{anthropic2024claude3}, enabling memory-guided decision-making. Prior work uses RL to select or refine skills within an agent’s repertoire. MemSkill~\citep{MemSkill} treats memory operations as learnable skills and trains a controller via RL to select appropriate memory behaviors. SkillRL~\citep{xia2026skillrl} jointly evolves the agent policy and a SkillBank by distilling successful trajectories into reusable strategies. Representing skills as executable code can further improve precision and reusability. PolySkill~\citep{yu2026polyskill} separates high-level skill abstractions from site-specific implementations to transfer skills across different web interfaces, while skill library-integrated GRPO~\citep{wang2025selfimproving} learns reusable action-sequence skills from long-horizon task chains, albeit with high training costs. While effective, these parametric approaches often incur expensive training and reduced generalization. In contrast, non-parametric methods improve agent behavior at inference time without updating model parameters. Skill-Pro~\citep{mi2026skillpro} learns reusable procedural skills from interaction experience, and MCE~\citep{ye2026metacontext} evolves skills through an \textit{agentic crossover} mechanism that recombines successful past behaviors. Similarly, our HORMA framework models memory management skills as non-parametric updates, enabling continual adaptation without modifying the underlying LLM.

\section{Preliminaries}
We consider long-horizon decision-making settings in which an LLM agent must solve a task specified by a natural language query $q$ through multi-step interaction with an environment. At each step $t \in [T]$, the agent receives an observation $o_t$ and produces an action $a_t$, forming an interaction trajectory $\mathbf{H}_{t-1} = (o_0, a_0, o_1, \dots, o_{t-1}, a_{t-1})$. The primary agent is modeled as an LLM-based policy $M_\theta$ with frozen parameters:
\begin{align}
M_\theta(a_t \mid o_t, \mathbf{H}_{t-1}, q; \mathcal{P}_{\text{main}}),
\end{align}
where $\mathcal{P}_{\text{main}}$ specifies the prompting context, including environment descriptions, tool specifications, output formats, and few-shot demonstrations.

\paragraph{The Context Bottleneck.}
While $q$ and $\mathcal{P}_{\text{main}}$ remain fixed, the interaction history $\mathbf{H}_t$ grows with trajectory length. Under a finite context window of size $W$, tokens exceeding the limit must be truncated, resulting in loss of long-range dependencies. Moreover, long histories introduce substantial computational overhead and information dilution, where task-relevant signals become increasingly obscured by irrelevant context. As trajectories grow, the agent must not only retain information under strict context budgets, but also organize and retrieve relevant information efficiently across long temporal horizons.

%\paragraph{Benefit of the Navigational Prior.} In this static context, $M_r$ can directly to target relevant dialogue turns without reading the entire history. This effectively transforms a linear search through $T$ turns into a targeted traversal of the hierarchy, ensuring that the primary agent $M_{\theta}$ receives a surgical context $\textbf{C}$ that is both token-efficient and semantically focused.

%\nikki{This section is too dense. Can we break it into two sections: move problem formulation / notations / setups to preliminaries, leaving this section solely focus on methodology and architecture design.}
\section{Hierarchical Organize-and-Retrieve Memory Agent}\label{sec:horma}
We present HORMA (\textbf{H}ierarchical \textbf{O}rganize-and-\textbf{R}etrieve \textbf{M}emory \textbf{A}gent), a framework that augments a primary LLM agent $M_\theta$ with an external working memory system. HORMA is motivated by the observation that memory construction and memory retrieval operate at fundamentally different temporal and functional scales. Memory construction shapes the long-term structure of stored information and induces delayed effects on downstream reasoning, whereas retrieval directly affects per-step inference quality on the execution path. We therefore explicitly decouple these processes into two specialized modules: a memory manager $M_m$ responsible for organizing information and a retrieval agent $M_r$ responsible for selecting task-relevant context.

Both modules are implemented as tool-using agents that interact with a shared hierarchical memory workspace exclusively through executable file-system operations and Bash tools. This shared grounded interface enables interpretable memory manipulation, explicit provenance tracking, and modular optimization of memory management and retrieval behaviors. The overall architecture is illustrated in Figure~\ref{fig:main} (1) \& (2).

\subsection{Memory-Augmented Agent Policy}
\label{sec:augmentMDP}

To address the limitations of growing interaction histories, HORMA externalizes working memory into a persistent hierarchical workspace that evolves alongside agent interaction. Rather than treating memory as a flat sequence of tokens, the workspace maintains structured and navigable representations of past experience, enabling memory construction and retrieval to operate independently from the primary agent's context window. We formalize the framework using a Memory-augmented Markov Decision Process (M-MDP)~\citep{zhou2025memento}, defined as $(\mathcal{S}, \mathcal{O}, \mathcal{A}, \mathcal{T}, \mathcal{R}, \mathcal{F})$. Here, $\mathcal{S}$, $\mathcal{O}$, and $\mathcal{A}$ denote state, observation, and action spaces; $\mathcal{T}$ defines environment dynamics; $\mathcal{R}$ provides a binary task reward $R_T \in \{0,1\}$ at the final step $T$, and $\mathcal{F}_t$ denotes the external memory state at time $t$.

The memory workspace $\mathcal{F}_t$ consists of structured files and directories (e.g., entity logs, event summaries, state trackers) that evolve through a memory transition operator:
\begin{align}
\mathcal{F}_{t+1} = \mathcal{T}_{\mathcal{F}}(\mathcal{F}_t, a_t, o_t).
\end{align}

Unlike interaction history stored directly in the context window, $\mathcal{F}_t$ persists externally and can scale with task complexity. While $\mathcal{T}_{\mathcal{F}}$ could be implemented using hand-crafted heuristics such as fixed summarization or rule-based file updates, such approaches are often brittle and fail to generalize across domains requiring complex management and retrieval strategies. To overcome these limitations, HORMA operationalizes this transition by framing memory construction as an agentic management task driven by a memory manager $M_m$, while decomposing downstream per-step action generation into localized retrieval and execution:
\[
\pi(a_t \mid o_t, \mathcal{F}_t, q)
=
M_r(\mathbf{C}_t \mid \mathcal{F}_t, q)
\;
M_\theta(a_t \mid o_t, \mathbf{C}_t, q; \mathcal{P}_{\text{main}}),
\]
where the retrieval module $M_r$ selects context $\mathbf{C}_t \subseteq \mathcal{F}_t$ to ground the primary agent $M_{\theta}$.

\paragraph{Generalization to Long-Horizon Conversations.}

Although formulated for interactive environments, the framework naturally extends to long-horizon conversational settings such as long-form QA and dialogue memory benchmarks~\citep{locomo, wu2025longmemeval}. In this setting, the interaction history becomes
\[
\mathbf{H} = (u_0, r_0, u_1, r_1, \dots, u_T, r_T),
\]
where $u_t$ and $r_t$ denote user and assistant turns. Standard approaches directly condition on the entire dialogue history:
\begin{align}
M_\theta(a \mid \mathbf{H}, q; \mathcal{P}_{\text{main}}),
\end{align}
which becomes increasingly inefficient as conversations grow. HORMA instead retrieves compact task-relevant context from external memory:
\[
\pi(a \mid \mathcal{F}, q)
=
M_r(\mathbf{C} \mid \mathcal{F}, q)
\;
M_\theta(a \mid \mathbf{C}, q; \mathcal{P}_{\text{main}}),
\]
enabling scalable reasoning over long conversational histories under strict context limits.

\subsection{The Grounded Workspace: Hierarchy and Provenance}

HORMA organizes memory $\mathcal{F}_t$ as a hierarchical file system rather than a flat memory buffer~\citep{memoryr1}. This design enables memory construction to operate over semantically meaningful structures while supporting efficient retrieval through directory navigation and localized search. For each interaction $(a_t, o_t)$ or dialogue turn $(u_t, r_t)$, the memory manager $M_m$ first archives the raw trajectory into a timestamped directory. It then selectively synthesizes structured notes based on task relevance. Each synthesized note stores compact task-relevant abstractions together with temporal metadata and references to underlying raw trajectories, enabling efficient retrieval without sacrificing provenance or recoverability.

%Each synthesized note contains (i) a concise high-density summary of task-relevant entities, events, or states; (ii) a file-path reference to the corresponding raw trajectory, and (iii) temporal and structural metadata. Unlike irreversible compression-based memory approaches, HORMA preserves linkage between synthesized notes and underlying trajectories, enabling efficient retrieval over compact abstractions while retaining provenance and access to fine-grained interaction details when necessary.

\subsection{Memory Management via Skill Evolution}

The memory manager $M_m$ maintains the workspace $\mathcal{F}_t$ by issuing executable file-system operations through Bash commands (e.g., \texttt{mkdir}, \texttt{nano}, \texttt{mv}). Its role is to transform raw trajectories into semantically organized memory structures that support long-horizon reasoning. Because memory construction operates over long temporal horizons and induces delayed structural effects on downstream reasoning, directly optimizing memory structure through sparse task rewards is highly unstable. We therefore treat memory construction as a structure induction problem. We initialize $M_m$ with a domain-agnostic prompt $\mathcal{P}_m^{(0)}$ that specifies high-level organizational principles such as entity tracking, event abstraction, and relation grouping. This initialization leverages the strong hierarchical reasoning and abstraction capabilities already exhibited by frontier LLMs.

\paragraph{Recursive Skill Refinement.}
To improve memory construction, we iteratively refine $\mathcal{P}_m$ using task trajectories. We identify failure modes by comparing performance using raw history $\mathbf{H}$ (unconstrained baseline) versus managed context $\mathbf{H'}$ (HORMA).
We categorize failures based on whether structured memory helps or hinders performance relative to unstructured history, enabling us to distinguish cases where memory construction removes information versus cases where it improves reasoning by filtering noise.
We define two contrastive subsets:
\begin{itemize}
    \item $\mathcal{D}_{\text{exo}}$ (Exogenous set): tasks where $\mathbf{H}$ succeeds but $\mathbf{H'}$ fails, indicating information loss during memory construction;
    \item $\mathcal{D}_{\text{end}}$ (Endogenous set): tasks where $\mathbf{H'}$ succeeds but $\mathbf{H}$ fails, indicating that structured memory mitigates issues such as hallucination or \textit{lost-in-the-middle} effects~\citep{liu2024lostmiddle}.
\end{itemize}

For each task, we generate natural language feedback via contrastive analysis:
\begin{equation}
\text{Feedback}_i = \text{LLM}(\text{Feedback Instruction}, \mathbf{H}, \mathbf{H'}).
\end{equation}

We aggregate feedback across tasks to iteratively refine the memory management policy with additional memory management skills:
\begin{equation}
\mathcal{P}_m^{(k+1)} = \text{LLM}(\text{Skill Augmentation Instruction},\mathcal{P}_m^{(k)}, \{\text{Feedback}_i\}_{i=1}^n),
\end{equation}
which can be viewed as a form of textual gradient descent~\citep{yuksekgonul2025optimizing}. This process yields a growing library of domain-specific memory management skills (exogenous and endogenous). The overall memory management pipeline is illustrated in Figure~\ref{fig:main} (1).

\subsection{Memory Retrieval via Reinforcement Learning}
\label{sec:rl_retrieval}

The retrieval agent $M_r$ navigates the memory workspace $\mathcal{F}_t$ to construct task-relevant context $\mathbf{C}_t$. Unlike similarity-based retrieval, which may retrieve temporally inconsistent or causally irrelevant information, $M_r$ exploits explicit structural signals such as directory hierarchy, temporal organization, and provenance metadata to efficiently locate relevant content, illustrated in Figure~\ref{fig:main} (2).

Retrieval decisions lie directly on the execution path and admit localized behavioral feedback, making retrieval naturally amenable to sequential policy optimization. We therefore formulate retrieval as a grounded decision-making process over executable file-system operations.

\paragraph{Grounded Action Space.}

The retrieval agent interacts with the workspace using Bash commands such as \texttt{ls}, \texttt{grep}, \texttt{cd}, and \texttt{cat}. We further augment the action space with two terminal actions:
\[
\{\texttt{select}, \texttt{done}\}.
\]

The \texttt{select} action adds verified content to the retrieved context $\mathbf{C}_t$, while \texttt{done} terminates retrieval once sufficient evidence has been collected. The primary agent acts conditioned on retrieved context:
\begin{align}
M_\theta(a_t \mid o_t, \mathbf{C}_t, q; \mathcal{P}_{\text{main}}).
\end{align}

This design improves efficiency by allowing retrieval to operate over compact structured notes rather than full trajectories while selectively expanding into raw interaction traces only when necessary.

\paragraph{RL-based Policy Optimization.}

To improve retrieval reliability under strict context constraints, we optimize the retrieval policy $M_r$ using Group Relative Policy Optimization (GRPO)~\citep{grpo}. 

\begin{comment}
    The objective combines a policy gradient term with a KL regularization penalty:
\begin{equation}
\mathcal{L}^{\text{GRPO}}(\phi)
=
\mathbb{E}[\cdots],
\end{equation}
where the policy gradient term encourages actions leading to higher-than-expected returns and the KL penalty constrains deviation from the reference policy to stabilize optimization.
\end{comment}

To encourage precise yet compact context construction, we define an evidence-grounded retrieval reward based on overlap between retrieved context $\mathbf{C}_t$ and ground-truth evidence $E$:
\begin{equation}
J(\mathbf{C}_t, E)
=
\frac{
|\mathbf{C}_t \cap E|
}{
|\mathbf{C}_t \cup E|
}.
\end{equation}

This reward encourages retrieval of relevant evidence while penalizing irrelevant or redundant context. Over time, the retrieval agent learns efficient navigation strategies such as hierarchical exploration, recovery from failed commands, and refinement of search trajectories, enabling robust and lightweight retrieval under limited context budgets.

\section{Experiments}\label{sec:experiment}
\subsection{Experimental Setup}

\paragraph{Benchmarks.} We evaluate our methods on three benchmarks: ALFWorld~\citep{alfworld}, LoCoMo~\citep{locomo}, and LongMemEval~\citep{wu2025longmemeval}. ALFWorld is an embodied interactive task benchmark, where we evaluate $134$ tasks across $6$ task categories. LoCoMo and LongMemEval are long-horizon conversational benchmarks designed to test memory construction from extended dialogue histories. LoCoMo contains $10$ conversations; we evaluate on $519$ question-answering instances drawn from $3$ conversations, while the remaining $7$ conversations are used to train the lightweight retrieval agent with Qwen 3.5 4B. We evaluate on LongMemEval with $367$ instances spanning diverse question types. For the main results (Table~\ref{tab:alfworld_parallel} and Table~\ref{tab:conversation_parallel}), we use Claude Sonnet 4.5 as the backbone model for reasoning, memory management, and retrieval across all compared methods. Variants using Qwen-based GRPO retrievers are analyzed separately in Section~\ref{sec:analysis} and Table~\ref{tab:main_results_structured}. Additional training details, experimental setups, and hyper-parameters are provided in Appendix~\ref{sec:appen_implement}.

\paragraph{Baselines and Metrics.} We compare against representative context management baselines, including static memory methods such as truncation, sliding window, and embedding-based similarity, as well as dynamic memory approaches such as ReSum~\citep{wu2025resum} and Acon~\citep{kang2025acon}. For ALFWorld, we additionally include context folding methods, where \textit{Fold} aggregates action-observation trajectories into their preceding reasoning steps within the ReAct framework~\citep{yao2023react}, as well as HIAGENT~\citep{hu2025hiagent}. For conversational benchmarks~\citep{locomo, wu2025longmemeval}, we further compare against external memory systems including A-MEM~\citep{xu2025amem} and Mem0~\citep{chhikara2025mem0}, along with an embedding-based similarity retrieval baseline built on our structured note representations. We evaluate both task performance and memory efficiency. On ALFWorld~\citep{alfworld}, we report success rate, along with the average number of interaction steps per task and the average input tokens per step, where total token usage is their product. On conversational benchmarks~\citep{locomo, wu2025longmemeval}, we report LLM-as-a-judge (L-J) scores using Claude Sonnet 4.5, with F1 scores provided in Appendix~\ref{appendix:setup}, as well as total input token usage per question-answering instance.

\begin{figure}[t]
     \centering
     % First Subfigure
     \begin{subfigure}[b]{0.48\textwidth}
         \centering
         \includegraphics[width=\columnwidth,keepaspectratio]{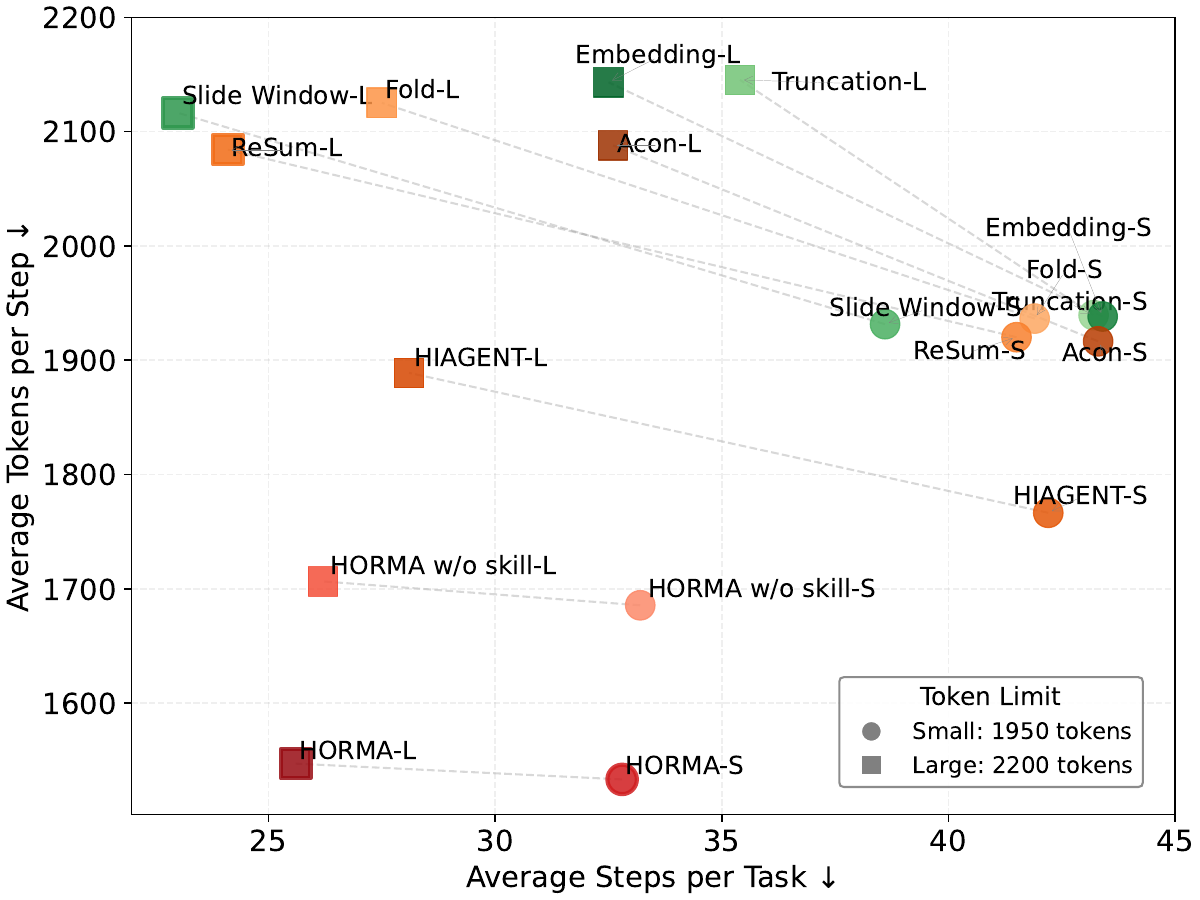}
         \caption{ALFWorld Pareto Efficiency}
         \label{fig:alfworld_token}
     \end{subfigure}
     % Second Subfigure
     \begin{subfigure}[b]{0.48\textwidth}
         \centering
         \includegraphics[width=\columnwidth,keepaspectratio]{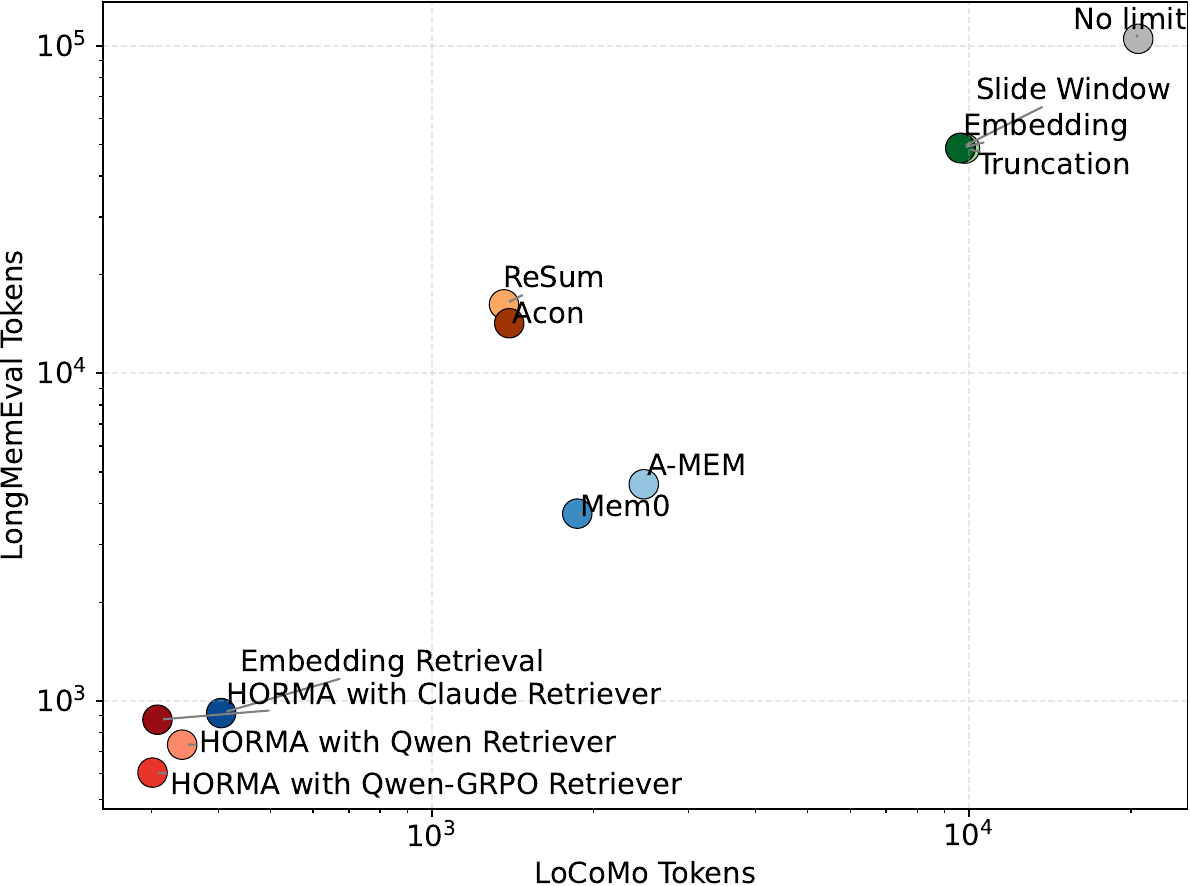}
         \caption{Conversation Benchmark Token Usage}
         \label{fig:conversation_token}
     \end{subfigure}
     
     \caption{Efficiency-Performance Trade-offs Across Benchmarks: (a) Comparison of average interaction steps versus tokens per step under Small (1950) and Large (2200) context limits; (b) Total input tokens consumed on LoCoMo (x-axis) and LongMemEval (y-axis) on a logarithmic scale.}
     \label{fig:main_token}
\end{figure}

\begin{table*}[t]
\centering
\caption{ALFWorld Performance with Claude Sonnet 4.5 as primary agent's backbone under different context windows: Small ($1950$ context input token limit) and Large ($2200$ context input token limit). We report Success Rate ($\%$). Subscripts denote differences relative to Truncation with {\color{mygreen}{improvement}} and {\color{red}{degradation}}. The best results are highlighted in \textbf{bold} and our methods are in \colorbox{lightblue!50}{blue}.}

\label{tab:alfworld_parallel}
\resizebox{\textwidth}{!}{%
\begingroup
\renewcommand{\arraystretch}{1.1}
\setlength{\tabcolsep}{3pt}

\begin{tabular}{lccccccc ccccccc}
\toprule
\multirow{2}{*}{\textbf{Method}}
& \multicolumn{7}{c}{\textbf{Small}}
& \multicolumn{7}{c}{\textbf{Large}} \\
\cmidrule(lr){2-8} \cmidrule(lr){9-15}

& \textbf{Pick} & \textbf{Clean} & \textbf{Heat} & \textbf{Cool} & \textbf{Look} & \textbf{Pick2} & \textbf{All}
& \textbf{Pick} & \textbf{Clean} & \textbf{Heat} & \textbf{Cool} & \textbf{Look} & \textbf{Pick2} & \textbf{All} \\

\midrule
\rowcolor{green!10}\multicolumn{15}{c}{\textbf{Internal Static Memory}}\\
\midrule

Truncation
& 50.0 & 3.2 & 13.0 & \textbf{23.8} & 0.0 & 17.6 & 17.9
& 91.7 & 0.0 & 47.8 & 4.8 & 55.6 & 11.8 & 34.3 \\

Slide Window
& \textbf{\gain{95.8}{45.8}} & \drop{0.0}{3.2} & \drop{0.0}{13.0} & \drop{0.0}{23.8} & \gain{66.7}{66.7} & \gain{23.5}{5.9} & \gain{29.1}{11.2}
& \gain{95.8}{4.1} & \gain{61.3}{61.3} & \gain{82.6}{34.8} & \gain{14.3}{9.5} & \textbf{\gain{83.3}{27.7}} & \textbf{\gain{88.2}{76.4}} & \gain{70.1}{35.8} \\

Embedding
& \gain{75.0}{25.0} & \drop{0.0}{3.2} & \drop{0.0}{13.0} & \drop{0.0}{23.8} & \gain{5.6}{5.6} & \drop{0.0}{17.6} & \drop{14.2}{3.7}
& \textbf{\gain{100.0}{8.3}} & \gain{3.2}{3.2} & \gain{50.0}{2.2} & \drop{0.0}{4.8} & \gain{77.8}{22.2} & \gain{35.3}{23.5} & \gain{42.5}{8.2} \\

\midrule
\rowcolor{orange!10}\multicolumn{15}{c}{\textbf{Internal Dynamic Memory}}\\
\midrule

Fold
& \gain{87.5}{37.5} & \drop{0.0}{3.2} & \drop{0.0}{13.0} & \drop{0.0}{23.8} & \gain{16.7}{16.7} & \drop{0.0}{17.6} & 17.9
& \gain{95.8}{4.1} & \gain{35.5}{35.5} & \textbf{\gain{87.0}{19.8}} & \drop{0.0}{4.8} & \gain{77.8}{22.2} & \gain{64.7}{52.9} & \gain{59.0}{24.7} \\

HIAGENT
& \gain{83.3}{33.3}
& \gain{6.5}{3.3}
& \drop{4.3}{8.7}
& \drop{4.8}{19.0}
& \gain{33.3}{33.3}
& \gain{35.3}{17.7}
& \gain{27.0}{9.1}
& \gain{95.8}{4.1}
& \gain{54.8}{54.8}
& \gain{78.3}{30.5}
& \gain{14.3}{9.5}
& \gain{77.8}{22.2}
& \gain{70.6}{58.8}
& \gain{64.9}{30.6}
\\

ReSum
& \gain{83.3}{33.3} & \drop{0.0}{3.2} & \drop{0.0}{13.0} & \drop{0.0}{23.8} & \gain{27.8}{27.8} & \drop{0.0}{17.6} & \gain{18.7}{0.8}
& \textbf{\gain{100.0}{8.3}} & \gain{67.7}{67.7} & \gain{78.3}{30.5} & \gain{33.3}{28.5} & \gain{72.2}{16.6} & \gain{76.5}{64.7} & \gain{71.6}{37.3} \\

Acon
& \gain{70.8}{20.8} & \drop{0.0}{3.2} & \drop{0.0}{13.0} & \drop{0.0}{23.8} & \gain{16.7}{16.7} & \drop{0.0}{17.6} & \drop{15.0}{2.9}
& \gain{95.8}{4.1} & \gain{6.5}{6.5} & 47.8 & \drop{0.0}{4.8} & \gain{77.8}{22.2} & \gain{41.2}{29.4} & \gain{42.5}{8.2} \\

\midrule
\rowcolor{purple!8}\multicolumn{15}{c}{\textbf{External Dynamic Memory}}\\
\midrule

%\rowcolor{lightblue!50}
%HORMA w/o skill & \gain{83.3}{33.3} & \gain{71.0}{67.8} & \drop{4.3}{8.7} & \drop{0.0}{23.8} & \gain{66.7}{66.7} & \gain{64.7}{47.1} & \gain{49.3}{31.4} & \drop{87.5}{4.2} & \gain{64.5}{64.5} & \drop{0.0}{47.8} & \gain{9.5}{4.7} & \textbf{\gain{83.3}{27.7}} & \gain{64.7}{52.9} & \gain{51.5}{17.2} \\

\rowcolor{lightblue!50}
HORMA
& \gain{87.5}{37.5} & \textbf{\gain{74.2}{71.0}} & \textbf{\gain{21.7}{8.7}} & \drop{4.8}{19.0} & \textbf{\gain{72.2}{72.2}} & \textbf{\gain{70.6}{53.0}} & \textbf{\gain{56.7}{38.8}}
& \drop{79.2}{12.5} & \textbf{\gain{83.9}{83.9}} & \gain{60.9}{13.1} & \textbf{\gain{95.2}{90.4}} & \gain{66.7}{11.1} & \gain{47.1}{35.3} & \textbf{\gain{73.9}{39.6}} \\

\bottomrule
\end{tabular}
\endgroup
}
\vspace{-1pt}
\end{table*}

\begin{table*}[t]
\centering
\caption{We report  LoCoMo ($10$K context  input token limit) and LongMemEval ($50$K context input token limit) performance with Claude Sonnet 4.5 as primary agent's backbone. We report \textbf{L-J} scores ($\uparrow$) on varying task types and Overall splits. Subscripts denote differences relative to Truncation baseline with {\color{mygreen}{improvement}} and {\color{red}{degradation}}. The best results except for \textit{No limit} are highlighted in \textbf{bold} and our methods are in \colorbox{lightblue!50}{blue}. Note that SS refers to single-session and KU refers to Knowledge Update.}
\label{tab:conversation_parallel}
\resizebox{\textwidth}{!}{%
\begingroup
\renewcommand{\arraystretch}{1.1}
\setlength{\tabcolsep}{3pt}

\begin{tabular}{lcccccccccc}
\toprule
\multirow{2}{*}{\textbf{Method}}
& \multicolumn{4}{c}{\textbf{LoCoMo}}
& \multicolumn{6}{c}{\textbf{LongMemEval}} \\
\cmidrule(lr){2-5} \cmidrule(lr){6-11}

& \textbf{Single-hop} & \textbf{Temporal} & \textbf{Adversarial} & \textbf{Overall}
& \textbf{KU} & \textbf{SS-assist} & \textbf{SS-prefer} & \textbf{SS-user} & \textbf{Temporal} & \textbf{Overall} \\

\midrule

No limit
& 78.1 & 66.4 & 1.5 & 55.9
& 37.2 & 21.4 & 3.3 & 20.0 & 14.3 & 20.4 \\

\midrule
\rowcolor{green!10}\multicolumn{11}{c}{\textbf{Internal Static Memory}}\\
\midrule
Truncation
& 47.8 & 30.8 & 0.7 & 32.2
& 62.8 & 37.5 & 6.7 & 42.9 & 17.3 & 34.1 \\

Slide Window
& \drop{39.9}{7.9}
& \gain{47.7}{16.9}
& 0.7
& \drop{31.4}{0.8}
& \drop{19.2}{43.6}
& \drop{8.9}{28.6}
& \drop{0.0}{6.7}
& \drop{25.7}{17.2}
& \drop{15.8}{1.5}
& \drop{16.1}{18.0} \\

Embedding
& \drop{47.5}{0.3}
& \gain{45.8}{15.0}
& \drop{0.0}{0.7}
& \gain{34.8}{2.6}
& \drop{37.2}{25.6}
& \drop{12.5}{25.0}
& \drop{3.3}{3.4}
& \drop{35.7}{7.2}
& 17.3
& \drop{23.2}{10.9} \\

\midrule
\rowcolor{orange!10}\multicolumn{11}{c}{\textbf{Internal Dynamic Memory}} \\
\midrule

ReSum 
& \drop{21.9}{25.9}
& \gain{32.7}{1.9}
& 0.7
& \drop{18.7}{13.5}
& \gain{60.3}{23.1}
& \drop{19.6}{17.9}
& \gain{33.3}{26.6}
& \gain{52.9}{10.0}
& \drop{15.8}{1.5}
& \gain{34.3}{0.2} \\

Acon 
& \drop{42.8}{5.0}
& \drop{26.2}{4.6}
& \textbf{\gain{16.4}{15.7}}
& \gain{32.6}{0.4}
& \gain{64.1}{1.3}
& \drop{26.8}{10.7}
& \textbf{\gain{40.0}{33.3}}
& \gain{61.4}{18.5}
& \gain{21.8}{4.5}
& \gain{40.6}{6.5} \\

\midrule
\rowcolor{purple!8}\multicolumn{11}{c}{\textbf{External Dynamic Memory}} \\
\midrule

A-MEM 
& \drop{39.6}{8.2}
& \gain{35.5}{4.7}
& \gain{7.5}{6.8}
& \drop{30.4}{1.8}
& \gain{64.1}{1.3}
& \gain{57.1}{19.6}
& \gain{33.3}{26.6}
& \gain{78.6}{35.7}
& \gain{32.3}{15.0}
& \gain{51.8}{17.7} \\

Mem0 
& \gain{64.7}{16.9}
& \drop{28.0}{2.8}
& \gain{11.2}{10.5}
& \gain{43.4}{11.2}
& \drop{42.3}{20.5}
& \gain{53.6}{16.1}
& \gain{10.0}{3.3}
& \gain{85.7}{42.8}
& \gain{25.6}{8.3}
& \gain{43.6}{9.5} \\

Embedding Retrieval
& \gain{57.2}{9.4}
& \gain{33.6}{2.8}
& 0.7
& \gain{37.8}{5.6}
& \drop{44.9}{17.9}
& \drop{21.4}{16.1}
& \gain{10.0}{3.3}
& \gain{51.4}{8.5}
& \gain{18.8}{1.5}
& \drop{30.2}{3.9} \\

%\rowcolor{lightblue!50} HORMA with Qwen retriever  & \drop{36.0}{11.8} & \drop{28.0}{2.8} & \gain{7.5}{6.8} & \drop{27.0}{5.2} & \drop{48.7}{14.1} & \drop{19.6}{17.9} & \drop{3.3}{3.4} & \drop{32.9}{10.0} & \gain{30.1}{12.8} & \drop{30.8}{3.3} \\

%\rowcolor{lightblue!50} HORMA with Qwen-GRPO retriever & \gain{61.2}{13.4} & \drop{30.0}{0.8} & \gain{12.7}{12.0} & \gain{42.2}{10.0} & \gain{83.3}{20.5} & \drop{35.7}{1.8} & \gain{36.7}{30.0} & \gain{82.9}{40.0} & \textbf{\gain{44.4}{27.1}} & \textbf{\gain{58.0}{23.9}} \\

\rowcolor{lightblue!50}
HORMA
& \textbf{\gain{70.5}{22.7}}
& \textbf{\gain{50.5}{19.7}}
& \gain{13.4}{12.7}
& \textbf{\gain{51.6}{19.4}}
& \textbf{\gain{89.7}{26.9}}
& \textbf{\gain{58.9}{21.4}}
& \gain{13.3}{6.6}
& \textbf{\gain{88.6}{45.7}}
& \textbf{\gain{33.1}{15.8}}
& \textbf{\gain{55.9}{21.8}} \\

\bottomrule
\end{tabular}
\endgroup
}
\vspace{-1pt}
\end{table*}

\begin{table*}[t]
\centering
\caption{Ablation of skill usage in management and RL (e.g., GRPO) in retrieval indicated by \checkmark/\xmark. We report task performance ($\uparrow$) (success rate for ALFWorld and L-J scores for conversational benchmarks) and the number of LLM retrieval calls ${N}_{\text{call}}$ ($\downarrow$) per interactive step or question-answering instance. The primary agent and memory manager use Claude Sonnet 4.5. Best results are in \textbf{bold}. Relative improvements over the no-skill, no-RL baseline are shown in {\color{mygreen}{green}}.}
\label{tab:main_results_structured}
\resizebox{\textwidth}{!}{%
\begin{tabular}{lcc|cc|cc|cc}
\toprule
\multirow{2}{*}{\textbf{Retriever}} 
& \multirow{2}{*}{\textbf{Skill}} 
& \multirow{2}{*}{\textbf{RL}} 
& \multicolumn{2}{c}{\textbf{ALFWorld (Large)}} 
& \multicolumn{2}{c}{\textbf{LoCoMo}} 
& \multicolumn{2}{c}{\textbf{LongMemEval}} \\
\cmidrule(lr){4-5} \cmidrule(lr){6-7} \cmidrule(lr){8-9}

& & 
& \textbf{Performance} & \textbf{${N}_{{call}}$} 
& \textbf{Performance} & \textbf{${N}_{{call}}$} 
& \textbf{Performance} & \textbf{${N}_{{call}}$} \\

\midrule

\multirow{2}{*}{Claude Sonnet 4.5}
& $\times$ & $\times$ & 51.5 &4.47 & 42.2 & 4.98 & 43.6 & 5.47 \\
& $\checkmark$ & $\times$ & \textbf{\gain{73.9}{22.4}} & \textbf{\improve{4.43}{0.04}} & \textbf{\gain{51.6}{9.4}} & \textbf{\improve{4.73}{0.25}} & \gain{55.9}{12.3} & \improve{5.26}{0.21} \\

\midrule

\multirow{3}{*}{Qwen 3.5 4B}
& $\times$ & $\times$ & 35.8 & 5.21 & 27.0 & 5.46 & 30.8 & 5.53 \\
& $\checkmark$ & $\times$ & \gain{40.3}{4.5} & \improve{5.08}{0.13} & \gain{32.6}{5.6} & \improve{5.33}{0.13} & \gain{40.6}{9.8} & \improve{5.41}{0.12} \\
& $\checkmark$ & $\checkmark$ & \gain{64.9}{29.1} & \improve{4.58}{0.63} & \gain{42.2}{15.2} & \improve{5.12}{0.54} & \textbf{\gain{58.0}{27.2}} & \textbf{\improve{5.22}{0.31}} \\

\bottomrule
\end{tabular}
}\vspace{-1pt}
\end{table*}

\begin{figure}[t]
     \centering
     % First Subfigure
     
     % Second Subfigure
     \begin{subfigure}[b]{0.32\textwidth}
         \centering
         \includegraphics[width=\columnwidth,keepaspectratio]{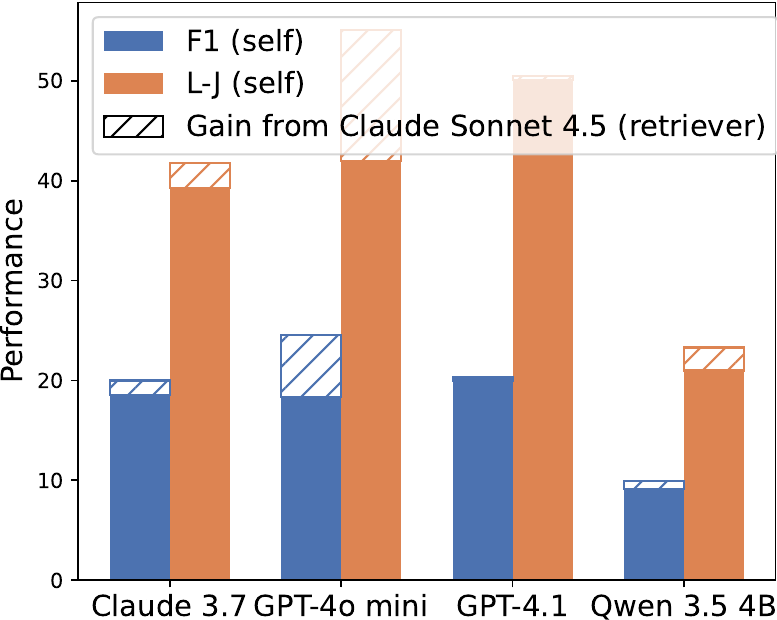}
         \caption{Memory construction bottleneck across varying memory manager}
         \label{fig:claude_retrieval}
     \end{subfigure}
     \hfill % Adds horizontal 
     \begin{subfigure}[b]{0.32\textwidth}
         \centering
         \includegraphics[width=\columnwidth,keepaspectratio]{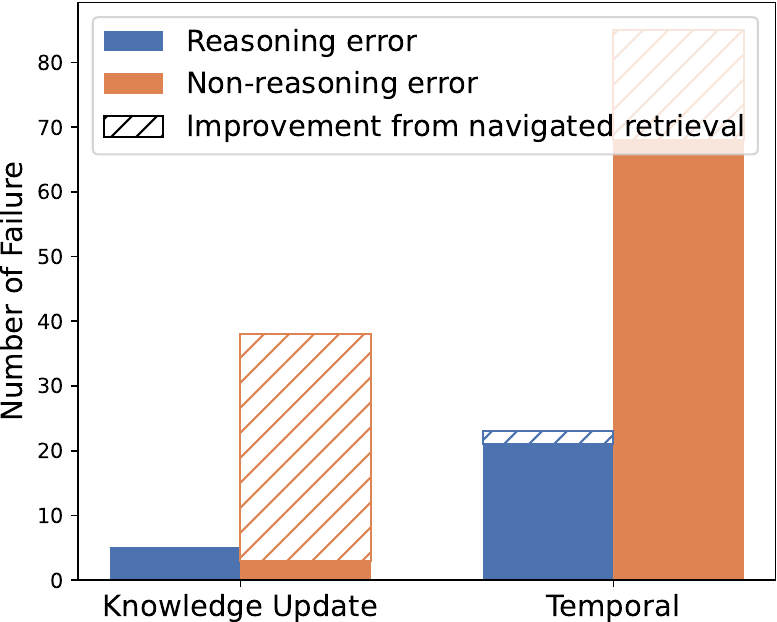}
         \caption{Error Attribution for temporally sensitive tasks}
         \label{fig:navigated_retrieval}
     \end{subfigure}
     \hfill % Adds horizontal spacing between figuresspacing between figures
     % Third Subfigure
     \begin{subfigure}[b]{0.34\textwidth}
         \centering
         \includegraphics[width=\columnwidth,keepaspectratio]{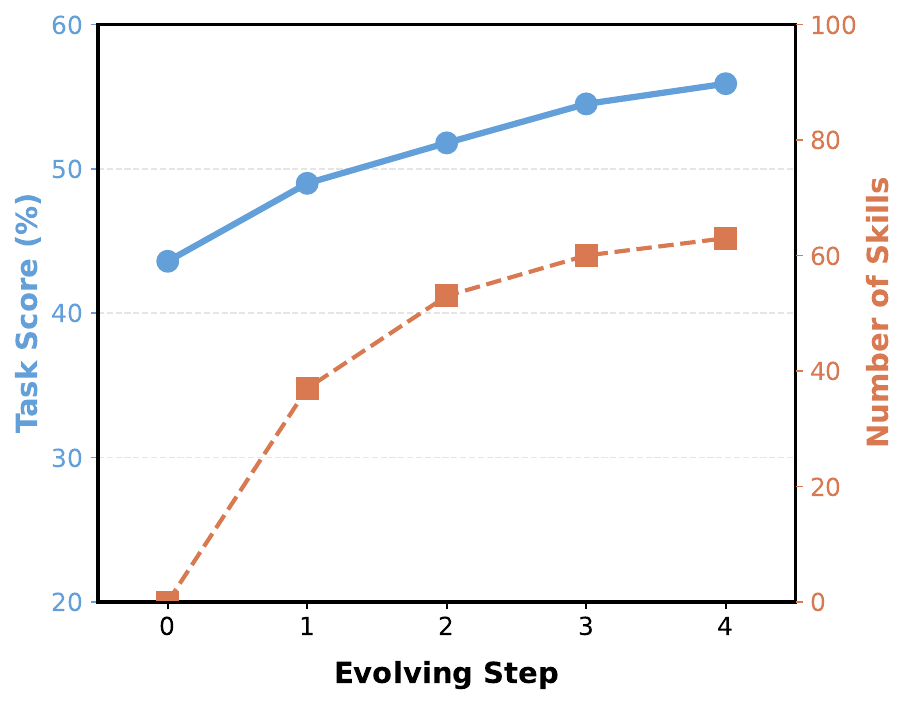}
         \caption{Skill library growth and scaling}
         \label{fig:longmemeval_skill}
     \end{subfigure}

     \caption{Analysis of Retrieval Reliability and Skill Acquisition: (a) Hashed bars as the gain when replacing native retrievers with a stronger retriever (Sonnet 4.5) in LoCoMo; (b) Comparison of failure modes for temporally sensitive tasks between similarity-based retrieval and HORMA’s navigated retrieval in LongMemEval; (c) The iterative expansion of the skill library over four refinement rounds in LongMemEval.}
     \label{fig:longmemeval_preliminary}
\end{figure}

\subsection{Main Results}
\paragraph{Interactive Benchmark.}
On ALFWorld, we evaluate all methods under two context window settings. We analyze Pareto efficiency in terms of interaction steps and input tokens per step in Figure~\ref{fig:alfworld_token}. Across methods, larger context windows generally reduce the number of interaction steps, as more information can be incorporated at each decision step. Our method consistently achieves both fewer interaction steps and lower token usage under both settings. Incorporating memory management skills further improves efficiency. Table~\ref{tab:alfworld_parallel} reports task success rates under both settings. Notably, the \textit{Fold} baseline underperforms sliding window despite preserving reasoning traces, likely because retained reasoning significantly increases context cost and reduces actionable capacity. Overall, HORMA (with skill) achieves the best performance, reaching 56.7\% and 73.9\% success rate under small and large context limits, respectively. These results demonstrate the effectiveness of structured memory and navigation-based retrieval under strict context constraints.

\paragraph{Conversational Benchmarks.}
On conversational benchmarks, we evaluate settings where input lengths exceed $20$K tokens in LoCoMo and $100$K tokens in LongMemEval, stressing long-context retrieval under extreme context constraints. We analyze token efficiency across all methods in Figure~\ref{fig:conversation_token}, including HORMA variants with different retrievers (Claude, Qwen, and Qwen-GRPO), as well as \textit{Embedding Retrieval} augmented with our agentic memory management. All HORMA variants and \textit{Embedding Retrieval} consistently operate within $1000$ tokens per query, demonstrating that the efficiency gains from our memory management are robust across retrieval backbones.

In downstream task performance in Table~\ref{tab:conversation_parallel}, replacing standard embedding as internal static memory with \textit{Embedding Retrieval} enhanced by our structured note representation improves results on both benchmarks, indicating that memory management alone strengthens even simple retrieval methods. 
Together with Figure~\ref{fig:conversation_token}, this suggests that our memory management generalizes beyond HORMA to alternative retrieval paradigms. We further report HORMA's performance using a single representative configuration of HORMA (Claude-based retriever) in Table~\ref{tab:conversation_parallel} while analyses of other HORMA variants are deferred to Section~\ref{sec:analysis}. HORMA on LoCoMo achieves performance approaching the no-context-limit setting and outperforms all baselines. On LongMemEval, where information dilution and \textit{lost-in-the-middle} effects are pronounced, several baselines even outperform no-limit counterparts due to implicit filtering of irrelevant context. HORMA follows this trend and achieves the best overall performance, demonstrating stronger robustness to long-context degradation.

\subsection{Analysis}\label{sec:analysis}
\paragraph{Ablation Studies.}

We conduct ablation studies to analyze the contributions of (i) self-evolving memory management skills and (ii) the lightweight retriever trained with RL post-training. Results are reported in Table~\ref{tab:main_results_structured}. HORMA without skill evolution already achieves competitive performance against the baselines in Tables~\ref{tab:alfworld_parallel} and~\ref{tab:conversation_parallel}, suggesting that the unified memory management prompt alone generalizes effectively across tasks. Incorporating self-evolving memory skills further improves task performance while consistently reducing the number of LLM retrieval calls across retriever backbones and benchmarks. Using Claude Sonnet 4.5 for all modules (primary agent, memory manager, and retriever) achieves the strongest performance on ALFWorld and LoCoMo, highlighting the effectiveness of structured command-based memory management and retrieval within a unified framework. We further evaluate an RL-trained lightweight retriever based on Qwen 3.5 4B, trained only on the LoCoMo training split. The learned retriever improves LoCoMo performance from 32.6\% to 42.2\% while also exhibiting strong zero-shot transfer to ALFWorld and LongMemEval, improving both task performance and retrieval efficiency by reducing unnecessary retrieval calls. Notably, despite being trained solely on conversational data, the retriever generalizes across domains without modification and achieves the best overall performance on LongMemEval (58\%), surpassing the Claude Sonnet 4.5-based retrieval configuration.

\paragraph{Memory Construction Bottleneck.} To test the hypothesis that effective memory construction requires high-level semantic reasoning~\citep{jiang2025hibench, son2026content}, we evaluate HORMA across varying backbones for both management and retrieval. As shown in Figure~\ref{fig:claude_retrieval}, while proprietary models establish a high performance ceiling, the smaller Qwen 3.5 4B lags significantly. Crucially, when we replace each model's native retriever with Claude Sonnet 4.5, the performance gains are non-uniform. The improvement for Qwen 3.5 remains marginal even when enhanced by the superior retrieval capabilities of Sonnet 4.5. This demonstrates that flawed memory organization cannot be compensated for by high-quality retrieval. If the manager fails to induce a coherent structure, even an optimal navigation policy is restricted by the deficiencies of the underlying workspace. These results empirically justify the necessity of high-capacity LLMs for the memory management role. The full cross-backbone performance on LoCoMo can be found in Table~\ref{tab:locomo_restructured} in Appendix.

\paragraph{Impact of Agentic Retrieval on Temporal Reasoning.}

Semantic similarity-based retrieval is widely used~\citep{xu2025amem, memoryr1}, with extensions such as two-stage retrieval~\citep{zhang2026memrl} and RL-based temporal-aware retrieval~\citep{memoryt1}. Table~\ref{tab:conversation_parallel} shows that HORMA consistently outperforms baselines across categories. We further analyze its impact on temporally sensitive tasks (e.g., Knowledge Update and Temporal) in LongMemEval. For a controlled comparison, we evaluate both \textit{Embedding Retrieval} and HORMA on top of the same memory management framework. We collect inference trajectories and categorize failure cases into (i) reasoning errors, where retrieved context is correct but not properly utilized, and (ii) non-reasoning errors, including temporal staleness and irrelevant retrieval. As shown in Figure~\ref{fig:navigated_retrieval}, HORMA significantly reduces non-reasoning errors compared to embedding-based retrieval, while both methods share the same primary agent and thus similar reasoning capability. This indicates that improvements primarily arise from more accurate retrieval of temporally relevant information.

\paragraph{Skill Library Growth.} Figure~\ref{fig:longmemeval_skill} illustrates the evolution of the skill library over multiple refinement rounds on LongMemEval. Starting from an empty set, the library expands to $63$ skills after four rounds. Task performance improves steadily as the skill set grows, indicating that accumulated skills provide increasingly effective guidance for memory management. Examples of agent-generated memory management skills are provided in Tables~\ref{tab:alfworld_skills},~\ref{tab:locomo_skills}, and~\ref{tab:longmemeval_skills} in Appendix~\ref{appen:skill}. Importantly, our framework starts from a domain-agnostic memory management prompt and progressively augments it with domain-specific skills learned through interaction, enabling adaptation across tasks.
\section{Conclusion}\label{sec:conclusion}
We introduced HORMA, a hierarchical organize-and-retrieve memory agent that decouples working memory into a high-level memory manager and a low-level retriever operating over a structured file-system workspace. By separating asynchronous memory organization from per-step retrieval, HORMA improves credit assignment, context efficiency, and scalability in long-horizon reasoning. The memory manager acquires organizational skills through recursive trajectory refinement, while the retriever is optimized with RL to navigate hierarchical memory via executable file-system operations. HORMA achieves consistent gains across three benchmarks while substantially reducing context usage and retrieval overhead, and the learned retrieval policy generalizes effectively across domains. Future work will extend the current evidence-based retrieval training framework to fully online interaction-driven learning while preserving HORMA’s modular design.

\bibliography{References}
\bibliographystyle{plainnat}

%%%%%%%%%%%%%%%%%%%%%%%%%%%%%%%%%%%%%%%%%%%%%%%%%%%%%%%%%%%%

\newpage
\appendix
\onecolumn
\newcolumntype{Y}{>{\raggedright\arraybackslash}X} % for left-aligned X

\section{Dataset and Experiment Setup}\label{appendix:setup}
Table~\ref{tab:benchmark_summary} summarizes the benchmarks and dataset splits used in our experiments. Additional details for each benchmark are provided below.
\begin{table}[ht]
\centering
\caption{Summary of benchmarks and data splits used in our experiments. We evaluate HORMA across embodied interaction (\mbox{ALFWorld}~\citep{alfworld}) and extended conversational settings (\mbox{LoCoMo}~\citep{locomo} and \mbox{LongMemEval}~\citep{wu2025longmemeval}) to test long-context reasoning, context efficiency, and out-of-distribution (OOD) generalization.}
\label{tab:benchmark_summary}
\small
\begin{tabular}{lcc p{5cm}}
\hline
\textbf{Benchmark / Dataset} & \textbf{\# Samples} & \textbf{Domain / Level} & \textbf{Notes} \\ \hline
\textbf{ALFWorld} & 134 tasks & Household interaction & Standard OOD split across 6 categories for evaluation; text-based planning \\
\textbf{LoCoMo (Train)} & 1,089 QA & Long-form dialogue & First 7 conversations; used to train the lightweight retriever \\
\textbf{LoCoMo (Test)} & 519 QA & Long-form dialogue & Remaining 3 conversations for evaluation \\
\textbf{LongMemEval} & 367 QA & Multi-session chat & Zero-shot OOD for evaluation \\ \hline
\end{tabular}
\end{table}
\subsection{ALFWorld}
\paragraph{Environment.} ALFWorld~\citep{alfworld} is a text-based interactive environment built on TextWorld, where agents operate in household settings via natural language. At each step, the agent issues an action and receives textual feedback from the environment. The goal is to complete high-level tasks (e.g., placing an object in a specified location) within a fixed horizon of $50$ steps.
Tasks often require long-horizon reasoning. Consequently, agents must perform effective planning, maintain subgoal structure, and explore systematically.
\paragraph{Tasks and Action Space.}
The action space of the interactive agent is summarized in Table~\ref{tab:alfworld_action_space}. We evaluate on the standard out-of-distribution (OOD) split consisting of $134$ tasks across six categories: Pick \& Place ($24$), Clean \& Place ($31$), Heat \& Place ($23$), Cool \& Place ($21$), Examine in Light ($18$), and Pick Two \& Place ($17$).
\paragraph{Context Budget.}
To study performance under constrained context windows, we first establish a near-upper-bound by deploying an unconstrained agent based on Claude Sonnet 4.5. This setting achieves a $97.0\%$ success rate. The per-episode token usage ranges from $1588$ to $3435$ tokens (median: $2172$, mean: $2207.8$).
Guided by this distribution, we define two context budgets, $1950$ and $2200$ tokens, to simulate realistic memory constraints while preserving performance sensitivity. These budgets enable controlled evaluation of how different methods trade off context efficiency and task success.

\begin{table}[ht]
\centering
\caption{Action space for ALFWorld~\citep{alfworld}, where (object) refers to manipulable objects and (receptacle) refers to receptacles or locations in the environment.}
\begin{tabularx}{\textwidth}{Y|Y}
\hline
\textbf{Action Type} & \textbf{Description} \\
\hline
\texttt{go to (receptacle)} & go to location \\
\texttt{take (object) from (receptacle)} & take object from receptacle or location \\
\texttt{use (object)} & use object \\
\texttt{move (object) to (receptacle)} & move object to receptacle or location \\
\texttt{open (receptacle)} & open receptacle \\
\texttt{close (receptacle)} & close receptacle \\
\texttt{toggle (object) (receptacle)} & turn object or receptacle  on\\
\texttt{clean (object) with (receptacle)} & clean object with receptacle \\
\texttt{heat (object) with (receptacle)} & heat object with receptacle \\
\texttt{cool (object) with (receptacle)} & cool object with receptacle  \\
\hline
\end{tabularx}\label{tab:alfworld_action_space}
\end{table}

\begin{table*}[t]
\centering
\caption{LoCoMo ($10$K context input token limit) and LongMemEval ($50$K context input token limit) performance with Claude Sonnet 4.5 as primary agent's backbone. We report \textbf{F1} scores ($\uparrow$) on varying task types and Overall splits. Subscripts denote differences relative to Truncation baseline with {\color{mygreen}{improvement}} and {\color{red}{degradation}}. The best results except for \textit{No limit} are highlighted in \textbf{bold} and our methods are in \colorbox{lightblue!50}{blue}. Note that SS refers to single-session and KU refers to Knowledge Update.}
\label{tab:conversation_f1}
\resizebox{\textwidth}{!}{%
\begingroup
\renewcommand{\arraystretch}{1.1}
\setlength{\tabcolsep}{3pt}

\begin{tabular}{lcccccccccc}
\toprule
\multirow{2}{*}{\textbf{Method}}
& \multicolumn{4}{c}{\textbf{LoCoMo}}
& \multicolumn{6}{c}{\textbf{LongMemEval}} \\
\cmidrule(lr){2-5} \cmidrule(lr){6-11}

& \textbf{Single-hop} & \textbf{Temporal} & \textbf{Adversarial} & \textbf{Overall}
& \textbf{KU} & \textbf{SS-assist} & \textbf{SS-prefer} & \textbf{SS-user} & \textbf{Temporal} & \textbf{Overall} \\

\midrule

No limit
& \gain{37.5}{17.0}
& \gain{17.9}{7.1}
& \gain{0.7}{0.2}
& \gain{24.0}{11.3}
& \drop{9.2}{12.2}
& \drop{6.2}{4.4}
& \drop{12.0}{1.5}
& \drop{4.5}{9.4}
& \drop{6.0}{0.9}
& \drop{6.9}{5.5} \\

\midrule
\rowcolor{green!10}\multicolumn{11}{c}{\textbf{Internal Static Memory}} \\
\midrule

Truncation
& 20.5 & 10.8 & 0.5 & 12.7
& 21.4 & 10.6 & 13.5 & 13.9 & 6.9 & 12.4 \\

Slide Window
& \drop{15.5}{5.0}
& \gain{13.3}{2.5}
& \gain{0.7}{0.2}
& \drop{11.2}{1.5}
& \drop{7.1}{14.3}
& \drop{5.4}{5.2}
& \drop{9.8}{3.7}
& \drop{7.5}{6.4}
& \gain{7.3}{0.4}
& \drop{7.2}{5.2} \\

Embedding
& \gain{20.2}{0.3}
& \gain{12.2}{1.4}
& 0.5
& \gain{13.4}{0.7}
& \drop{8.8}{12.6}
& \drop{7.7}{2.9}
& \drop{11.0}{2.5}
& \drop{7.9}{6.0}
& \gain{7.8}{0.9}
& \drop{8.2}{4.2} \\

\midrule
\rowcolor{orange!10}\multicolumn{11}{c}{\textbf{Internal Dynamic Memory}} \\
\midrule

ReSum
& \drop{11.7}{8.8}
& \drop{10.6}{0.2}
& \drop{0.4}{0.1}
& \drop{8.6}{4.1}
& \gain{23.2}{1.8}
& \gain{12.7}{2.1}
& \drop{12.5}{1.0}
& \gain{30.0}{16.1}
& \gain{9.1}{2.2}
& \gain{16.9}{4.5} \\

Acon
& \gain{23.2}{2.7}
& \drop{10.1}{0.7}
& \textbf{\gain{9.7}{9.2}}
& \drop{10.4}{2.3}
& \gain{24.5}{3.1}
& \gain{13.3}{2.7}
& \gain{15.2}{1.7}
& \gain{32.7}{18.8}
& \gain{12.6}{5.7}
& \gain{25.3}{12.9} \\

\midrule
\rowcolor{purple!8}\multicolumn{11}{c}{\textbf{External Dynamic Memory}} \\
\midrule

A-MEM
& \gain{23.0}{2.5}
& \gain{12.2}{1.4}
& \gain{1.9}{1.4}
& \gain{13.2}{0.5}
& \gain{21.7}{0.3}
& \gain{15.5}{4.9}
& \textbf{\gain{18.8}{5.3}}
& \gain{30.0}{16.1}
& \gain{10.6}{3.7}
& \gain{18.6}{6.2} \\

Mem0
& \gain{25.6}{5.1}
& \drop{10.1}{0.7}
& \gain{2.6}{2.1}
& \gain{16.1}{3.4}
& \drop{16.3}{5.1}
& \gain{14.6}{4.0}
& \gain{14.2}{0.7}
& \gain{35.3}{21.4}
& \gain{9.0}{2.1}
& \gain{15.7}{3.3} \\

Embedding Retrieval
& \gain{26.5}{6.0}
& \gain{15.7}{4.9}
& \drop{0.4}{0.1}
& \gain{15.0}{2.3}
& \drop{13.7}{7.7}
& \drop{6.6}{4.0}
& \gain{14.2}{0.7}
& \gain{17.7}{3.8}
& \gain{7.3}{0.4}
& \drop{11.2}{1.2} \\

%\rowcolor{lightblue!50}HORMA with Qwen retriever & \drop{15.8}{4.7} & \drop{10.7}{0.1} & \gain{6.2}{5.7} & \drop{10.2}{2.5} & \drop{13.7}{7.7} & \gain{11.5}{0.9} & \drop{10.1}{3.4} & \drop{8.1}{5.8} & \gain{14.6}{7.7}& \gain{22.7}{10.3} \\

%\rowcolor{lightblue!50}HORMA with Qwen-GRPO retriever & \gain{29.8}{9.3} & \gain{11.9}{1.1} & \gain{7.4}{6.9} & \gain{16.7}{4.0} & \gain{32.7}{11.3} & \gain{12.5}{1.9} & \drop{11.7}{1.8} & \textbf{\gain{62.8}{48.9}} & \textbf{\gain{19.0}{12.1}} & \textbf{\gain{31.7}{19.3}} \\

\rowcolor{lightblue!50}HORMA
& \textbf{\gain{32.3}{11.8}}
& \textbf{\gain{16.6}{5.8}}
& {\gain{7.6}{7.1}}
& \textbf{\gain{21.4}{8.7}}
& \textbf{\gain{34.6}{13.2}}
& \textbf{\gain{15.8}{5.2}}
& \drop{10.7}{2.8}
& \textbf{{\gain{56.9}{43.0}}}
& \textbf{{\gain{17.3}{10.4}}}
& \textbf{{\gain{30.2}{17.8}}} \\

\bottomrule
\end{tabular}
\endgroup
}
\vspace{3pt}
\end{table*}

\subsection{LoCoMo}
LoCoMo~\citep{locomo} consists of long, multi-session dialogues designed to evaluate memory and reasoning over extended contexts. The benchmark contains $10$ conversations, each comprising $19$–$32$ sessions, where each session includes multiple dialogue turns.
We focus on question-answering tasks that probe short-term and compositional reasoning, specifically from three categories: (i) Single-hop, where answers are grounded in a single session; (ii) Temporal, which require reasoning over temporal relationships and tracking time-dependent cues across sessions; and (iii) Adversarial, which are constructed to induce incorrect responses, requiring the agent to recognize unanswerable or misleading queries.

\paragraph{Data Split.}
We use the first $7$ conversations for training, yielding $1089$ question-answering instances, and evaluate on $519$ instances from the remaining $3$ conversations. This split prevents exposure to evaluation dialogues during training, ensuring a strict separation of conversational context.

\paragraph{Context Budget.}
Following the setup in ALFWorld~\citep{alfworld}, we first analyze the unconstrained setting. The average token length per instance exceeds $20$K tokens, reflecting the long-horizon nature of the benchmark. Based on this distribution, we impose a $10$K token context budget to evaluate different methods under constrained memory, enabling systematic comparison of context efficiency and reasoning performance.

\subsection{LongMemEval}
To assess out-of-domain (OOD) generalization, we adopt LongMemEval~\citep{wu2025longmemeval}, a benchmark designed to evaluate memory retrieval and reasoning over long, multi-session conversations. We evaluate on $367$ question-answering instances spanning diverse task types.
The benchmark includes: (1) Single-session-user ($70$) and Single-session-assistant ($56$), which test the ability to recall information introduced by the user or assistant within a single session; (2) Single-session-preference ($30$), which evaluates whether the model can leverage user-specific information to generate personalized responses; (3) Knowledge Update (KU) ($78$), which requires tracking changes in user state and updating stored memory accordingly; and (4) Temporal Reasoning (TR) ($133$), which involves reasoning over both metadata timestamps and explicit temporal references.

\paragraph{Context Budget.}
We analyze the unconstrained setting by deploying an answering agent based on Claude Sonnet 4.5. The average context length exceeds $100$K tokens per instance, reflecting the substantial memory demands of the benchmark. To enable controlled evaluation, we impose a $50$K token context budget, allowing us to compare methods in terms of both context efficiency and reasoning performance under realistic constraints.

\section{Implementation Details}
\label{sec:appen_implement}

In Table~\ref{tab:alfworld_parallel} and Table~\ref{tab:conversation_parallel}, all methods, including HORMA, use Claude Sonnet 4.5 as the primary agent to ensure fair comparison. Unless otherwise specified, HORMA additionally employs Claude Sonnet 4.5 as both the memory manager and retrieval agent. We further investigate lightweight open-source retrievers based on Qwen 3.5 4B, including RL post-training, as reported in Table~\ref{tab:main_results_structured}. Table~\ref{table:grpo_hyper} summarizes the hyper-parameters used for GRPO training. Our lightweight retrieval agent is trained on the LoCoMo training split, which consists of $7$ conversations and $1089$ question-answering tasks. We evaluate the trained agent across all benchmarks to assess cross-domain generalization. The evidence set used for computing the evidence-grounded reward (defined in Section~\ref{sec:rl_retrieval}) is directly derived from the original dataset~\citep{locomo}. We also study the impact of varying memory management and retrieval backbones in Table~\ref{tab:locomo_restructured}. All RL-based retriever training experiments were conducted using 4 NVIDIA H200 GPUs. Inference with closed-source models was performed on a CPU-based virtual machine equipped with an Intel Xeon Platinum 8488C processor with 48 physical cores (96 logical CPUs).

\vspace{10pt}
\begin{table}[t]
\centering
\footnotesize
\renewcommand{\arraystretch}{0.85}
\setlength{\tabcolsep}{2pt}
\caption{Cross-Backbone Performance on LoCoMo. We evaluate the impact of different LLM backbones for memory management and retrieval strategies.}
\label{tab:locomo_restructured}
\vskip 0.05in
\resizebox{0.45\textwidth}{!}{%
\begin{tabular}{ll|cc|c}
\toprule
Management & Retrieval & F1 & L-J & Tokens \\
\midrule

\multirow{4}{*}{Claude Sonnet 4.5}
 & Claude Sonnet 4.5 & 21.4 & 51.6 & 308.1 \\
 & Claude 3.7 Sonnet & 18.5 & 49.5 & 1052.7  \\
 & GPT-4.1 & 21.2 & 50.7 & 547.0  \\
 & GPT-4o mini & 20.7 & 41.2 & 505.4  \\
\midrule

\multirow{2}{*}{Claude 3.7 Sonnet}
 & Claude 3.7 Sonnet & 18.5 & 39.3 & 619.6  \\
 & Claude Sonnet 4.5 & 20.0 & 41.8 & 475.8  \\
\midrule

\multirow{2}{*}{GPT-4.1}
 & GPT-4.1 & 20.0 & 50.1 & 262.3  \\
 & Claude Sonnet 4.5 & 20.3 & 50.5 & 267.9  \\
\midrule

\multirow{2}{*}{GPT-4o mini}
 & GPT-4o mini & 18.3 & 42.0 & 204.0  \\
 & Claude Sonnet 4.5 & 24.6 & 55.1 & 287.9  \\
\midrule

\multirow{2}{*}{Qwen 3.5 4B}
 & Qwen 3.5 4B & 9.1 & 21.0 & 301.7  \\
 & Claude Sonnet 4.5 &  9.9& 23.3 & 287.9 \\

\bottomrule
\end{tabular}
}
\end{table}

\begin{table}[htbp]
\centering

\caption{Hyper-parameters for GRPO training on memory retrieval agent.}

\vspace{0.3cm}
\label{table:grpo_hyper} 
\begin{tabular}{ p{4cm} p{3cm}  }
  \toprule
    Hyper-parameters           & Values                 \\ 
  \midrule
   
    Learning rate                                           & $ 10^{-6}$  \\ 
    Batch size                                           & 64  \\ 
    Mini-batch size                                           & 16  \\
    Rollouts per task $G$ & 8  \\

    Clip ratio $G$ & 0.2  \\

    KL coefficient & 0.01\\
     Entropy coefficient & 0.001\\

      Rollout temperature & 1.0\\
      
      top-p  & 0.95\\

    Max response length & 2048  \\

    Max interaction turn & 10 \\
    
  \bottomrule
\end{tabular} 
\end{table}

\section{Memory Management Skill Examples}
\label{appen:skill}In Table~\ref{tab:alfworld_skills} through Table~\ref{tab:longmemeval_skills}, we provide representative examples of endogenous and exogenous memory management skills across all three benchmarks, detailing their IDs, titles, and underlying principles. The specific logic for inducing these skills is defined by the contrastive discovery protocols in Prompt~\ref{box:endogenous_discovery} (Endogenous) and Prompt~\ref{box:exogenous_discovery} (Exogenous). These prompts are further structured by the JSON output requirements defined in the skill fields specification in Prompt~\ref{box:skill_specification} .

These skills are generated to refine domain-specific memory management guidelines via the contrastive analysis of past trajectories. For example, the
end\_002 Failed-Action Loop Detection in Table~\ref{tab:alfworld_skills} addresses the \textit{Nothing happens} environment response, a failure mode unique to ALFWorld. We also observe notable consistency in skill acquisition across similar domains. Because LoCoMo and LongMemEval are both multi-session conversational benchmarks, they share identical endogenous skills such as end\_001 Temporal Precision Anchoring and end\_002 Verbatim Quote Preservation in Table~\ref{tab:locomo_skills} and Table~\ref{tab:longmemeval_skills}.

\begin{table}[ht]
\centering
\caption{Examples of generated endogenous and exogenous skills for ALFWorld~\citep{alfworld}.}
\label{tab:alfworld_skills}
\small
% "l" for ID, "p{3.5cm}" for Skill Title, and "X" for wrapping the Principle column
\begin{tabularx}{\textwidth}{l p{3.5cm} X}
\toprule
\textbf{ID} & \textbf{Skill Title} & \textbf{Principle (Actionable Pattern)} \\ \midrule
\multicolumn{3}{l}{\textit{\textbf{Endogenous Skills}}} \\
end\_001 & Goal-Object Spatial Pinning & When a target object (CD, keychain, watch) is discovered, immediately write its exact location to memory with high-relevance tagging. Long raw traces bury the critical 'cd 3 on desk 2' observation under dozens of failed navigation steps; structured memory surfaces it as a top-priority fact. \\ \addlinespace
end\_002 & Failed-Action Loop Detection & Detect and break out of repeated failed actions (e.g., 'go to X' returning 'Nothing happens' multiple consecutive times) by consulting memory for alternative unexplored locations. Raw history makes it hard to recognize the loop pattern; structured memory flags 'last 5 actions: same target, all failed'. \\ \addlinespace
end\_003 & Multi-Step Plan Persistence & Store intermediate goals (found first CD, now need second CD) in memory so the agent can resume a multi-object task after 20+ intervening steps. Raw history forces re-parsing the entire trace to recall 'I already have keychain 1, need keychain 2'; structured memory keeps a 'task\_progress' counter. \\ \midrule

\multicolumn{3}{l}{\textit{\textbf{Exogenous Skills}}} \\
exo\_001 & Preserve Object Identity Chains & External memory must maintain explicit object identifiers (e.g., 'mug 1', 'potato 2', 'soapbar 3') across all write operations, never abstracting to generic 'a mug' or 'the potato'. Each memory file should include a persistent object registry mapping IDs to last-known locations and states.\\ \addlinespace
exo\_002 & Receptacle State Persistence & Memory must track open/closed state of all containers (cabinets, drawers, fridge, microwave) and their contents at each step. Retrieval queries about 'where is X' must return both the object's location and whether that location is accessible (open) or requires an open action first.\\ \addlinespace
exo\_003 & Appliance Interaction Failure Detection & Heating and cooling tasks require toggling appliances (microwave, fridge). Memory must explicitly track toggle attempts and their outcomes, distinguishing between 'item placed in appliance', 'appliance toggled successfully', and 'item retrieved after heating/cooling'. Generic 'heated X' claims without verified toggle success cause false completion. \\ \bottomrule
\end{tabularx}
\end{table}

\newpage

\begin{table}[ht]
\centering
\caption{Examples of generated endogenous and exogenous skills for LoCoMo~\citep{locomo}.}
\label{tab:locomo_skills}
\small
% "l" for ID, "p{3.5cm}" for Skill Title, and "X" for wrapping the Principle column
\begin{tabularx}{\textwidth}{l p{3.5cm} X}
\toprule
\textbf{ID} & \textbf{Skill Title} & \textbf{Principle (Actionable Pattern)} \\ \midrule
\multicolumn{3}{l}{\textit{\textbf{Endogenous Skills}}} \\
end\_001 & Temporal Precision Anchoring & Embed exact dates, session timestamps, and relative-time calculations (e.g., 'yesterday' relative to Session 6 = July 5, 2023) in every event note to prevent temporal hallucinations common when models must infer dates across thousands of conversational turns without explicit anchors. \\ \addlinespace
end\_002 & Verbatim Quote Preservation & Store exact conversational quotes with speaker attribution and turn IDs in structured notes, preventing paraphrasing drift and enabling fact verification. Long raw histories cause models to conflate similar statements or invent plausible-sounding but incorrect paraphrases.\\ \addlinespace
end\_003 & Retrieval Path Redundancy & Critical facts must be indexed through multiple retrieval paths (by person, by date, by topic, by keyword) with consistent content or reliable cross-references. Single-path storage causes retrieval failures when query uses alternate framing or when primary file is missed by grep. \\ \midrule

\multicolumn{3}{l}{\textit{\textbf{Exogenous Skills}}} \\
exo\_001 & Retrieval Path Redundancy & Critical facts must be indexed through multiple retrieval paths (by person, by date, by topic, by keyword) with consistent content or reliable cross-references. Single-path storage causes retrieval failures when query uses alternate framing or when primary file is missed by grep. \\ \addlinespace
exo\_002 & Negative Evidence Handling & External memory systems must explicitly mark when evidence is genuinely absent vs. when it exists but wasn't retrieved, preventing false 'not mentioned' responses. Baseline's long-context overload causes it to incorrectly claim absence for facts present in the trace.\\ \addlinespace
exo\_003 & Causal Chain Preservation & When dialogue describes reasoning, motivations, or cause-effect relationships, the full logical chain must be stored, not just the conclusion. Questions asking 'why' or 'what makes' require intermediate steps that summaries often discard.\\
\bottomrule
\end{tabularx}
\end{table}

\newpage

\begin{table}[ht]
\centering
\caption{Examples of generated endogenous and exogenous skills for LongMemEval~\citep{wu2025longmemeval}.}
\label{tab:longmemeval_skills}
\small
% "l" for ID, "p{3.5cm}" for Skill Title, and "X" for wrapping the Principle column
\begin{tabularx}{\textwidth}{l p{3.5cm} X}
\toprule
\textbf{ID} & \textbf{Skill Title} & \textbf{Principle (Actionable Pattern)} \\ \midrule
\multicolumn{3}{l}{\textit{\textbf{Endogenous Skills}}} \\
end\_001 & Temporal Precision Anchoring & Externalize every event with exact dates, session timestamps, and relative-time calculations in structured notes, preventing the LLM from hallucinating or losing temporal facts buried in long raw traces. Baseline fails to locate dates like 'two months ago' or 'last Wednesday' in 100k+ token histories; llm file retrieves pre-computed anchors from dedicated temporal files.\\ \addlinespace
end\_002 & Verbatim Quote Preservation & Anchor every fact to exact conversation quotes with turn IDs, preventing hallucination when baseline fabricates plausible-sounding but incorrect details from vague memory of long traces. Llm file's 'Original Quote' sections with source turn numbers enable verification and eliminate confabulation. \\ \addlinespace
end\_003 & Entity-Centric File Organization & Organize memory by question-relevant entities (people, places, events, devices) in dedicated folders, enabling targeted retrieval instead of forcing the LLM to scan entire conversation for scattered mentions. Baseline drowns in irrelevant dialogue; llm file retrieves only pertinent entity files.\\ \midrule

\multicolumn{3}{l}{\textit{\textbf{Exogenous Skills}}} \\
exo\_001 & Encode Negative Evidence Explicitly & When a question asks about information NOT mentioned (iPad purchase, cow purchase from Peter), the memory system must explicitly record what was NOT discussed to enable 'information not provided' answers. Absence-of-evidence is invisible in selective extraction pipelines.\\ \addlinespace
exo\_002 & Cross-Reference Comparative Facts & Questions comparing two events ('which happened first: A or B') require both facts stored with comparable timestamps in retrievable proximity. Siloed notes (A in purchases.md, B in events.md) prevent the temporal comparison even when both facts were recorded.\\ \addlinespace
exo\_003 & Implement Retrieval Coverage Diagnostics & When retrieval returns partial results, the agent cannot distinguish 'fact not in memory' from 'fact in memory but not retrieved'. Diagnostic metadata (files searched, match scores, coverage gaps) enables the agent to request broader search or admit uncertainty appropriately.\\
\bottomrule
\end{tabularx}
\end{table}

\newpage

\section{Prompt Template}

\begin{planbox}[box:alfworld_prompt]{ALFWorld Interactive Prompt}
You are an intelligent agent in a household environment and your target is to perform actions (aligns with subgoal) to complete the task goal. At the beginning of your interactions, you will be given the detailed description of the current environment and your goal to accomplish.

For each of your turn, you will be given the useful context for current turn. Your output must strictly follow this format:

<your next action>.\\

The available actions are: 
\begin{enumerate}
    \item go to (receptacle
    \item take (object) from (receptacle)
    \item use (object)
    \item move (object) to (receptacle)
    \item open (receptacle)
    \item close (receptacle)
    \item toggle (object) (receptacle)
    \item  clean (object) with (receptacle)
    \item heat (object) with (receptacle)
    \item cool (object) with (receptacle)
    \item think: (your thought)
\end{enumerate}
, where (object) refers to manipulable objects and (receptacle) refers to receptacles or locations in the environment. \\

\begin{enumerate}
    \item If the environment output: Nothing happens, that means the previous action is invalid and you should try more options.
    \item You can only hold one object at a time. Before taking a new object, make sure you have placed down any object you are currently holding.
    \item  You should not assume or anticipate the feedback.
    \item  Even if you have planned multiple steps ahead, you should only execute one action at a time, which aligns with subgoal.
    \item Do not proceed with any further exploration or actions until you receive the feedback from the environment after your action.
    \item Do not keep thinking.
\end{enumerate}

Your response should use one of the following formats:

<your next action> 

think: <your thoughts>.

Here are two examples.

<example1>

<example2>

Here is the task <task>.

\end{planbox}
\newpage

\begin{planbox}[box:locomo_prompt]{LoCoMo Answer Prompt}
Based on the following conversation, answer the question with a short, precise answer. Use the date/time information derived or concluded from the conversation sessions if it is temporal question. Pay close attention to who said what.

Conversation: \{context\}

Question: \{question\}

Answer:"""
\end{planbox}

\begin{planbox}[box:longmemeval_prompt]{LongMemEval Answer Prompt}
Based on the following conversation, answer the question with a short, precise answer. Use the date/time information derived or concluded from the conversation sessions if it is temporal question. Pay close attention to who said what.

Conversation: \{context\}

Question: \{question\}

Answer:"""

\end{planbox}

\begin{planbox}[box:endogenous_discovery]{Dynamic Endogenous Skill Discovery}
You are an expert in contrastive analysis of interactive agents with different memory setups.

EXPERIMENTAL SETTING (READ CAREFULLY)
We compare two runs of the \textbf{same} task\_id.
\begin{itemize}
    \item BASELINE simulates \textbf{no practical context limit}:  the agent's policy model can use the \textbf{full raw interaction trace} (steps or dialogue turns as exported) in context (no llm\_file / external folder memory). 
    \item OUR METHOD: an \textbf{external memory folder} updated and read via \textbf{Bash} (files, notes, summaries). The live policy prompt is not carrying the entire raw trace.
\end{itemize}

\textbf{Outcome here}: OUR METHOD succeeded. BASELINE failed.
{original memory management prompt}

TASK

Task ID: {ID}

Task Category: {task}

Initial Observation: {initial observation}

Success Trajectory: {OUR METHOD's trajectory}

Failed Trajectory: {BASELINE's trajectory}

YOUR JOB — ENDOGENOUS SKILLS (PRIMARY HYPOTHESIS)

Your main hypothesis is that baseline failed because super-long raw history in context
causes hallucination, lost-in-the-middle effects, or attention dilution, not because the task is impossible. Endogenous skills are advantages of compressed /structured external memory that avoid relying on an ever-growing raw trace.

For each skill, tie it to why raw-history overload hurts and how OUR METHOD sidesteps it.
Do not reduce this to generic “better planning” unless you anchor it in context-length /memory-structure mechanisms.

<Skill Fields Specification>
\end{planbox}
\newpage

\begin{planbox}[box:exogenous_discovery]{Dynamic Exogenous Skill Discovery}
You are an expert in contrastive analysis of interactive agents with different memory setups.

EXPERIMENTAL SETTING (READ CAREFULLY)
We compare two runs of the \textbf{same} task\_id.
\begin{itemize}
    \item BASELINE simulates \textbf{no practical context limit}:  the agent's policy model can use the \textbf{full raw interaction trace} (steps or dialogue turns as exported) in context (no llm\_file / external folder memory). 
    \item OUR METHOD: an \textbf{external memory folder} updated and read via \textbf{Bash} (files, notes, summaries). The live policy prompt is not carrying the entire raw trace.
\end{itemize}

\textbf{Outcome here}: OUR METHOD failed. BASELINE succeeded.
{original memory management prompt}

TASK

Task ID: {ID}

Task Category: {task}

Initial Observation: {initial observation}

Success Trajectory: {BASELINE's trajectory}

Failed Trajectory: {OUR METHOD's trajectory}

YOUR JOB — EXOGENOUS SKILLS (PRIMARY HYPOTHESIS)

Your main hypothesis is information loss in the Bash-managed llm file pipeline, e.g, facts never written, overwritten summaries, wrong file read, grep misses, truncation. Exogenous skills are concrete gaps or failure modes in external memory that this contrast exposes. If the trace includes Bash/file/memory tool strings, use them as evidence.

<Skill Fields Specification>
\end{planbox}
\begin{planbox}[box:skill_specification]{Skill Fields Specification}
\textbf{Skill fields (every skill):}
\begin{itemize}
    \item \textbf{skill\_id}: end\_001, end\_002,... in order (unique within this JSON).
    \item \textbf{title}: 3-5 words only (no trailing punctuation).
    \item \textbf{principle}: 1-2 sentences stating the core transferable idea.
    \item \textbf{memory\_prompt\_improvement}: 1-3 sentences - how to change the memory-writer / extraction prompt or schema so future runs better encode this skill (or "" if not applicable).
\end{itemize}

Return only valid JSON (no markdown fences, no extra text) with this schema:
\begin{center}
\begin{minipage}{0.9\linewidth}
\texttt{\{} \\
\texttt{~~"task\_id": ...,} \\
\texttt{~~"task\_category": "...",} \\
\texttt{~~"comparison\_type": "endogenous" or "exogenous",} \\
\texttt{~~"winning\_method": "ours\_llm\_file" or "baseline\_full\_raw\_history",} \\
\texttt{~~"losing\_method": "baseline\_full\_raw\_history" or "ours\_llm\_file",} \\
\texttt{~~"hypothesis\_alignment": ...,} \\
\texttt{~~"divergence": \{} \\
\texttt{~~~~"first\_divergence\_step": <int or null>,} \\
\texttt{~~~~"summary": "<when trajectories diverge>"} \\
\texttt{~~\},} \\
\texttt{~~"skills": [ \{} \\
\texttt{~~~~"skill\_id": "end\_001" or "exo\_001",} \\
\texttt{~~~~"title": "<exactly three to five words>",} \\
\texttt{~~~~"principle": "<one or two sentences>",} \\
\texttt{~~~~"evidence": ...,} \\
\texttt{~~~~"how\_to\_reinforce/acquire": ...,} \\
\texttt{~~~~"memory\_prompt\_improvement": ...} \\
\texttt{~~\} ],} \\
\texttt{~~"root\_cause\_summary": "<2-4 sentences>"} \\
\texttt{\}}
\end{minipage}
\end{center}
\end{planbox}

\begin{comment}
\begin{planbox}[box:skill_specification]{Skill Fields Specification}
\begin{lstlisting}[breaklines=true, breakatwhitespace=true]
**Skill fields (every skill):**
- skill_id: end_001, end_002,... in order (unique within this JSON).
- title: 3-5 words only (no trailing punctuation).
- principle: 1-2 sentences stating the core transferable idea.
- memory_prompt_improvement: 1-3 sentences - how to change the memory-writer / extraction prompt or schema so future runs better encode this skill (or "" if not applicable).

Return only valid JSON (no markdown fences, no extra text) with this schema:
{
  "task_id": ...,
  "task_category": "...",
  "comparison_type": "endogenous" or "exogenous",
  "winning_method": "ours_llm_file" or "baseline_full_raw_history,
  "losing_method": "baseline_full_raw_history" or "ours_llm_file",
  "hypothesis_alignment": ...
},
  "divergence": {
    "first_divergence_step": <int or null>,
    "summary": "<when trajectories diverge; link to memory vs raw-history stress>"
  },
  "skills": [
    {
      "skill_id": "end_001" or "exd_001",
      "title": "<exactly three to five words>",
      "principle": "<one or two sentences>",
      "evidence": ...,
      "how_to_reinforce/acquire": ...,
      "memory_prompt_improvement": ...
    }
  ],
  "root_cause_summary": "<2-4 sentences>"

\end{lstlisting}
\end{planbox}
\end{comment}

\begin{planbox}[box:llm_judge]{LLM Judge Prompt for Conversational Tasks}

You are an expert judge evaluating and labeling an answer to a question as 'CORRECT' or 'WRONG'.
You will be given the following data:
\begin{itemize}
    \item a question
    \item a gold (ground truth) answer
    \item a generated answer, which you will score as CORRECT or WRONG.
\end{itemize}

The point of the question is to ask about something based on two users' conversations. The gold answer will usually be a concise and short answer that includes the referenced topic, for example:

Question: Do you remember what I got the last time I went to Hawaii?

Gold answer: A shell necklace\\

The generated answer might be longer, but you should be generous with your grading, as long as it touches on the same topic as the gold answer, it should be counted as CORRECT.

For time-related questions, the gold answer will be a specific date, month, or year. The generated answer might include relative references (e.g., last Tuesday), but you should be generous if it refers to the same time period as the gold answer, mark it CORRECT, even if the format differs (e.g., "May 7th" vs. "7 May").\\

Now it's time for the real question:

Question: {question}

Gold answer: {gold\_answer}

Generated answer: {generated\_answer}\\

First, provide a short (one sentence) explanation of your reasoning, then finish with CORRECT or WRONG.

Do NOT include both CORRECT and WRONG in your response, or it will break the evaluation script.

Return the label in JSON format with the key as "label".

\end{planbox}

\section{Memory Management: Hierarchical Workspace Examples}
The following excerpts illustrate the structured memory representations generated by the memory manager using Claude Sonnet 4.5 as the backbone. These examples, drawn from the LoCoMo benchmark, demonstrate how raw dialogue is transformed into a navigable, file-centric hierarchy that preserves temporal anchors and provenance.
\begin{tcolorbox}[title= People/Caroline.md]
LGBTQ Support Group Attendance
\begin{itemize}
    \item[-] Attended an LGBTQ support group on May 7, 2023 (the day before the conversation) (D1:3)
    \item[-] Found the experience "powerful" (D1:3)
    \item[-] Was inspired by transgender stories shared at the group (D1:5)
    \item[-] The group made her feel accepted and gave her courage (D1:7)
\end{itemize}

Career Plans
\begin{itemize}
    \item[-] Interested in counseling or mental health work (D1:11)
    \item[-] Wants to support people with similar issues (D1:11)
    \item[-] Planning to continue education and explore career options (D1:9)
\end{itemize}

\end{tcolorbox}

\begin{tcolorbox}[title= Dates/May2023.md]
May 7, 2023
\begin{itemize}
    \item[-] Caroline attended an LGBTQ support group (D1:3)
    \item[-] She found the experience "powerful" (D1:3)
\end{itemize}

May 8, 2023
\begin{itemize}
    \item[-] Caroline and Melanie had their first conversation session at 1:56 pm (Session 1)
    \item[-] Caroline shared about her experience at the LGBTQ support group from the previous day (D1:3)
\end{itemize}
\end{tcolorbox}

\begin{tcolorbox}[title= Topics/LGBTQ/support group.md]
Caroline's Attendance
\begin{itemize}
    \item[-] Caroline attended the group on May 7, 2023 (the day before their conversation) (D1:3)
    \item[-] "I went to a LGBTQ support group yesterday and it was so powerful." (D1:3)
\end{itemize}

Impact on Caroline
\begin{itemize}
    \item[-] She found transgender stories shared at the group inspiring (D1:5)
    \item[-] "The support group has made me feel accepted and given me courage to embrace myself." (D1:7)
    \item[-] The experience may have influenced her interest in counseling or mental health work (D1:11)

\end{itemize}
\end{tcolorbox}

\newpage

%%%%%%%%%%%%%%%%%%%%%%%%%%%%%%%%%%%%%%%%%%%%%%%%%%%%%%%%%%%%

\end{document}